\documentclass[11pt,a4paper]{article}
\usepackage[hyperref]{emnlp2020}
\usepackage{times}
\usepackage{latexsym}

\usepackage{setspace}
\usepackage{times}
\usepackage{latexsym}
\usepackage{graphicx}
\usepackage{url}
\usepackage{multirow}
\usepackage{algorithm}
\usepackage{algorithmic}
\usepackage{amsmath}
\usepackage{enumerate}
\usepackage{balance}
\usepackage{enumitem}
\usepackage{array}

\usepackage{booktabs}
\usepackage{bm}
\usepackage{amssymb}
\usepackage{amsfonts}
\usepackage{mathtools}
\usepackage{comment}
\usepackage{subfigure}

\usepackage{xcolor}
\definecolor{ao}{rgb}{0.0, 0.5, 0.0}
\newcommand{\error}[1]{\textcolor{red}{#1}}
\newcommand{\style}[1]{\textcolor{blue}{#1}}

\usepackage{soul}
\DeclareRobustCommand{\hlcyan}[1]{{\sethlcolor{cyan}\hl{#1}}}

\usepackage{hyperref}
\usepackage{url}

\newcolumntype{L}[1]{>{\raggedright\arraybackslash}p{#1}}
\newcolumntype{R}[1]{>{\raggedleft\arraybackslash}p{#1}}
\newcolumntype{C}[1]{>{\centering\let\newline\\\arraybackslash\hspace{0pt}}m{#1}}
\usepackage{microtype}

\aclfinalcopy % Uncomment this line for the final submission
\title{
Data-to-Text Generation with Style Imitation
}

\author{Shuai Lin$^{1,2}$,~~ Wentao Wang$^{2}$,~~ Zichao Yang$^{2}$,~~ Xiaodan Liang$^1$\thanks{~~corresponding authors}, ~~ Frank F. Xu$^{2}$ \\ 
{\bf Eric P. Xing}$^{2,3}${\bf ,}~~ {\bf Zhiting Hu}$^{2,4*}$\\
{\small $^1$Sun Yat-sen University,~ $^2$Carnegie Mellon University,~ $^3$Petuum Inc.,~ $^4$UC San Diego} \\
{\small\texttt{\{shuailin97,xdliang328,zhitinghu\}@gmail.com, \{zichaoy,fangzhex,epxing\}@cs.cmu.edu}}
}
\date{}

\begin{document}
\maketitle
\begin{abstract}
%Control over textual characteristics in natural language generation is an essential demand in real-world applications. Data-to-Text generation is one of the applications, where recent neural approaches have mostly focused on improving content fidelity while lacking the control over writing styles (e.g., sentence structures, word choices). More traditional systems use templates to determine the realization of text. 
Recent neural approaches to data-to-text generation have mostly focused on improving content fidelity while lacking explicit control over writing styles (e.g., word choices, sentence structures). More traditional systems use templates to determine the realization of text. Yet manual or automatic construction of high-quality templates is difficult, and a template acting as hard constraints could harm content fidelity when it does not match the record perfectly. We study a new way of stylistic control by using existing sentences as ``soft'' templates. That is, the model learns to imitate the writing style of any given exemplar sentence, with automatic adaptions to faithfully describe the content record. 
%has been studied mostly for improving content fidelity. However, many practical situations require to go beyond mere content correctness and further control the writing style, e.g., to convey brand personalities in writing advertising copies, or embody different authorial styles when producing sports commentaries. We study this new setting by proposing data-to-text generation with style imitation, which aims to generate a record description that imitates the writing style of any given reference text. Style imitation generalizes classical template-based generation by allowing to leverage existing raw texts as a comprehensive pool of writing styles. 
The problem is challenging due to the lack of parallel data. We develop a neural approach that includes a hybrid attention-copy mechanism, learns with weak supervisions, and is enhanced with a new content coverage constraint. 
%We further derive two large style imitation datasets for the study. 
We conduct experiments in restaurants and sports domains.
Results show our approach achieves stronger performance than a range of comparison methods. Our approach balances well between content fidelity and style control given exemplars that match the records to varying degrees.\footnote{Data and code are publicly available at \url{https://github.com/ha-lins/DTG-SI}}
% Our data and code are publicly available at \url{https://github.com/ha-lins/Text-Generation-with-Style-Imitation}.

%For example, controlling the writing style of data-to-text generation enables to write advertising copies that convey 

%writing ad copy to convey brand personality 

% {\bf
% Controllable text generation has posed an important role in nature language generation field due to its high practical usage. Recent efforts have made impressive progress in generating or editing sentences with given textual attributes (e.g., sentiment). This paper studies a new practical setting of \emph{text content manipulation}. Given a structured record, such as \emph{(PLAYER: Lebron, POINTS: 20, ASSISTS: 10)}, and a style sentence for reference, such as \emph{Kobe easily dropped 30 points}, the task aims at generating a sentence that \emph{accurately} describes the content in the record, with the designated writing style (e.g., wording, transitions, and order). The problem combines the characteristics of data-to-text
% generation and style transfer, and is challenging due to the trade-off between manipulating the text minimally and ensuring fidelity to the structured data. We derive two datasets from the data-to-text task as our testbed, and develop a neural method with unsupervised competing objectives and explicit content coverage constraint. Quantitative and human evaluations show the superiority of our approach over competitive methods including a template-based baseline and prior approaches designed for style transfer.
% }
\end{abstract}

\section{Introduction}

Recent years have seen remarkable progress in \emph{neural} natural language generation to produce well-formed coherent text~\citep{sutskever2014sequence,vaswani2017attention}. Yet, controllability over various text properties, as an essential demand to ensure the utility of generations in real-world applications, has not attained the same level of advancement.
%to ensure utility of the generation in designated situations, controllability over various text properties is an essential demand. 
Data-to-text generation is one of such applications with ubiquitous practical use, in which natural language text is generated to describe a given data record such as a box score of a sports player or an infobox table of a restaurant. 

\begin{figure*}[t]
\centering
\includegraphics[width=0.99\linewidth]{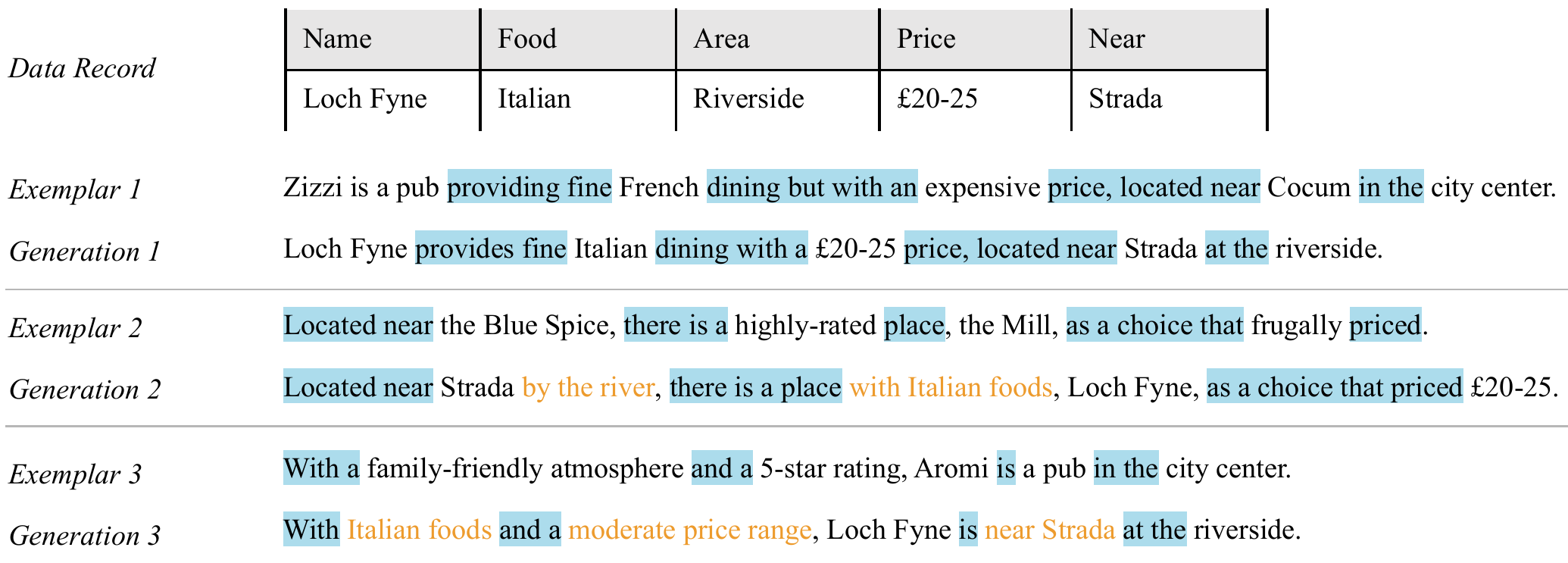}
% \vspace{-8pt}
\caption{An example of generating sentences that describe the data record and imitate the style of given exemplar sentences (i.e., soft templates). The generations \emph{adaptively} inherit the structural and phrasing characteristics (highlighted with \hlcyan{cyan} boxes) of the exemplars. For instance, exemplar~2 does not match the record content perfectly (e.g., it does not describe the food). The generation adapts the structure to add ``with Italian foods''. All such automatic adaptions are highlighted in {\color{orange}orange}. Note that the word ``providing'' in exemplar~1 is also adapted to ``provides'' for grammar correction.
}
%\vspace{-0.4cm}
\label{fig:task}
\end{figure*}

Though current data-to-text neural approaches with encoder-decoder models could produce fluent text with high fidelity to content (``what to say''), they largely lack control over the writing style, such as sentence structures and word choices (``how to say''). Many efforts have been made to promote the overall diversity in data-to-text generation through, e.g., latent variables~\cite{ye2020variational} or customized model architectures~\cite{jagfeld-etal-2018-sequence,deriu-cieliebak-2018-syntactic}. Yet fine-grained style manipulation is not permitted. 
This contrasts with the traditional text generation systems which separate content planning and surface realization~\cite{reiter1997building}, and usually determine the realization with explicit templates~\cite{kukich-1983-design,mcroy-etal-2000-yag} or based on syntactic grammars~\cite{robin1996empirically,power2003generating}. 

Controlling writing style with ``hard'' templates could suffer from unscalable template creation and lack of generation flexibility. Though previous work~\cite{wiseman2018learning,dou2018data2text,angeli2010simple} has enabled automatic template extraction, the templates usually act as hard constraints and could harm the content fidelity of generations when the template does not exactly match the content in a record.

In this paper, we study a new way of stylistic control in data-to-text generation by using any existing sentences as ``soft'' templates. That is, we learn to \emph{imitate} the writing style of a given exemplar sentence. The goal is two-fold: to generate text that not only faithfully describes all content in the record, but also inherits as many of the exemplar's stylistic characteristics as possible (Figure~\ref{fig:task}). The new paradigm sidesteps the restrictions with traditional dedicated templates and allows us to use arbitrary exemplar sentences that could be describing distinct content. As shown in Figure~\ref{fig:task}, the model automatically adapts the soft templates to varying extents based on how well they match the record, and precisely expresses the desired content.

To this end, we develop a neural approach that balances well between content fidelity and style imitation. 
%This paper develops a neural approach that learns adaptive style imitation to balance well between content fidelity and style embodiment. 
A key learning challenge is the lack of parallel data, i.e., triples of (record, exemplar sentence, target description). Instead, we usually only have access to abundant record-description pairs\footnote{This highlights the difference from the recent retrieval-and-generation work~\citep[e.g.,][]{hashimoto2018retrieve,weston2018retrieve,cao2018retrieve,peng2019text} which focuses only on content fidelity and thus is a supervised learning problem given the record-description pairs.}. 
%incorporates retrieved auxiliary sentences primarily for content fidelity, and thus is in essence a supervised learning problem given record-description pairs.
%The added control over writing style requires new learning solutions. 
The proposed approach learns with rich weak supervisions derived from the record-description pairs. Architecture-wise, we develop a hybrid attention-copy mechanism that offers differentiated treatments of the content and style sources. Further, based on the structural nature of data records, we devise a new content coverage constraint for the balanced embodiment of both content and style in the generation.
%and an explicit content coverage constraint for balanced embodiment of both content and style in the generation. \hzt{More model details?}

We conduct empirical studies on corpora from two domains, including restaurant recommendation~\cite{dusek2019e2e} and NBA reports~\cite{wiseman2017challenges}. 
%For empirical studies and future research, we also introduce two large datasets in different domains, which are derived from the existing restaurant recommendations~\cite{dusek2019e2e} and NBA reports~\cite{wiseman2017challenges} corpora, respectively.
Experiments show our models strongly improves over a diverse set of comparison methods in terms of both automatic and human evaluations. In particular, given exemplar sentences that match data records to varying degrees, our approach retains a good content-style balance.

\section{Related Work}
\label{sec:related}

\paragraph{Data-to-Text Generation}
Many efforts have been made to improve the fidelity of generated text to the record content, through sophisticated neural architectures~\citep{wiseman2017challenges,gehrmann2018end,puduppully-etal-2019-data,iso-etal-2019-learning}, hybrid retrieval and generation~\citep{hashimoto2018retrieve,weston2018retrieve,cao2018retrieve,pandey2018exemplar,peng2019text}, and others.
% Generating text conditioning on structured input has been widely studied in recent work~\citep[etc]{wen2015semantically,lebret2016neural,yang2016reference,wiseman2017challenges}. Those methods are based on neural sequence to sequence models and trained with supervised data. 
% This line of work has focused primarily on generating more accurate descriptions of the given data, while it has not studied the problem of controlling the writing style of outputs. 
These approaches do not have the additional goal of style control as ours, and usually perform supervised learning based on record-description pairs.
%Our task takes a step forward to \emph{simultaneously} describing desired content and controlling stylistic characteristics. Furthermore, our task is challenging due to its unsupervised setting in practice.
Traditional data-to-text generation systems implement a pipeline architecture consisting of separate components, including content planning, sentence planning, and surface realization~\cite[e.g.,][]{reiter1997building,kukich-1983-design,mcroy-etal-2000-yag,kondadadi-etal-2013-statistical}. Recent work~\cite{wiseman2018learning} integrates the template use in a more end-to-end neural model. Rather than treating templates as hard constraints as in the previous work, we study the new setting of using existing sentences as exemplars, allowing the model to adaptively imitate the style while ensuring content fidelity.

%Beyond generating text from scratch, there is another line of work that first retrieves an analogous sentence and then rewrites it to express desired information~\citep{hashimoto2018retrieve,weston2018retrieve,li2018delete,guu2018generating}.
%For example, \citet{weston2018retrieve} used the framework to generate response in dialogues, while \citet{hashimoto2018retrieve} studied programming code generation. The goal of the work is to manifest useful information from neighbors, usually in a supervised context, without aiming at controlling writing characteristics, and thus has fundamentally different assumptions to ours.

\paragraph{Text Style Transfer}
There has been growing interest in text style transfer~\citep[etc]{hu2017controllable,shen2017style,yang2018unsupervised,subramanian2019multiple} which assumes an existing sentence of certain content, and modifies single
or multiple textual attributes (e.g., sentiment) of the sentence without changing the content. Our problem differs in important ways in that we assume the abstract writing style is encoded in an exemplar sentence and attempts to modify its concrete content to express the new information in a structured record (we thus can call our setting \emph{text content rewriting}). The different settings can lead to different application scenarios in practice, and pose varying technical challenges. In particular, though the recent style transfer research~\citep{subramanian2019multiple,logeswaran2018content} has controlled multiple categorical attributes which are largely independent or loosely correlated to each other, a data record in our task, in comparison, can contain a varying number of fields, have many possible values, and are structurally coupled.
% A model must understand the content structure, and minimally yet
% sufficiently manipulate the reference sentence by rewriting, adding, or deleting text portions, with necessary polishing for grammatical correctness and fluency. 
% We name the problem text content manipulation. 
Our empirical studies (sec~\ref{sec:exp}) show the recent models designed for style transfer fail to perform well on the problem under study. 
We also note recent work of syntactically-controlled paraphrase generation based on either constituency parse~\cite{iyyer2018adversarial} or reference sentences~\cite{chen2019controllable}. The problem nature of data-to-text generation in this work leads to a solution with very different architectures and learning approaches.

\paragraph{Controlled Generation without Parallel Data}
Controlling different aspects (e.g., content, style, discourse structures) in text generation requires grasping the intrinsic mapping between the aspects and the surface text. The lack of parallel data often poses challenges in learning the mapping, making it necessary to incorporate other forms of experiences (supervisions)~\citep{hu2020learning}. For example, the style transfer work~\cite{hu2017controllable,shen2017style,yang2018unsupervised} used auxiliary models such as attribute classifiers and language models for supervision signals. \citet{tang2019target} learned guided conversation flow using standard conversation data combined with logical control. \citet{tan2020summarizing} created weak supervision labels from knowledge bases for aspect-based summarization. This work devises competing training objectives based on common record-description pairs. Joint optimization of the competing objectives drives the model to learn desired behaviors.

\section{The Task: Data-to-Text Generation with Style Imitation}
\label{sec:task}

For clarity, we first formally describe the problem of data-to-text generation with style imitation. We also establish the key notations used in the paper.

% We first establish the notations used in the rest of the paper. We then formally define the problem of unsupervised text content manipulation by introducing two competitive desiderata.
% We then illustrate the generality of the task via comparing with two existing tasks.

Consider a data record $\bm{x}$ which consists of a set of \emph{fields} and their values (e.g., field ``Food'' and its value ``Italian'' in Figure~\ref{fig:task}). Note that different records can include different fields. For example, the field ``Customer Rating'' is included in some records but not the one in Figure~\ref{fig:task}. Data-to-text generation aims to produce a sentence to describe the content in the record. We are additionally given an exemplar sentence $\bm{y}_{e}$ which could be describing distinct content in the same domain. 
%For a valid definition of style imitation, we assume the fields described in $\bm{y}_{sty}$ overlap at least partly with those in the record $\bm{x}$. 
The goal of the task is thus to generate a new sentence $\bm{y}$ that achieves {\bf (1) content fidelity} by describing the content in $\bm{x}$ accurately and completely, and {\bf (2) style embodiment} by retaining as much of the writing style (e.g., sentence structure, word choice, etc) of $\bm{y}_{e}$ as possible. 
%The task is unsupervised as there is no ground-truth sentence for training. 

A solution to the problem is required to balance well between the two objectives, by adaptively rewriting necessary portions of the reference $\bm{y}_{e}$ to express the desired content in a correct and fluent way, while at the same time editing $\bm{y}_{e}$ to a minimum extent to inherit its style. The demand for adaptive trade-off necessitates developing learning approaches for flexible imitation and generation.
%to go beyond traditional rule-based approaches 

To the best of our knowledge, there is no large data containing the desired $(\bm{x}, \bm{y}_{e}, \bm{y})$ \emph{triples} for supervised learning. Instead, we often only have access to \emph{pairs} of record and its description which was originally written without following any designated style. In the next section, we develop a neural approach that learns style imitation given only the paired data.

\begin{figure*}[t]
\centering
  \includegraphics[width=\linewidth]{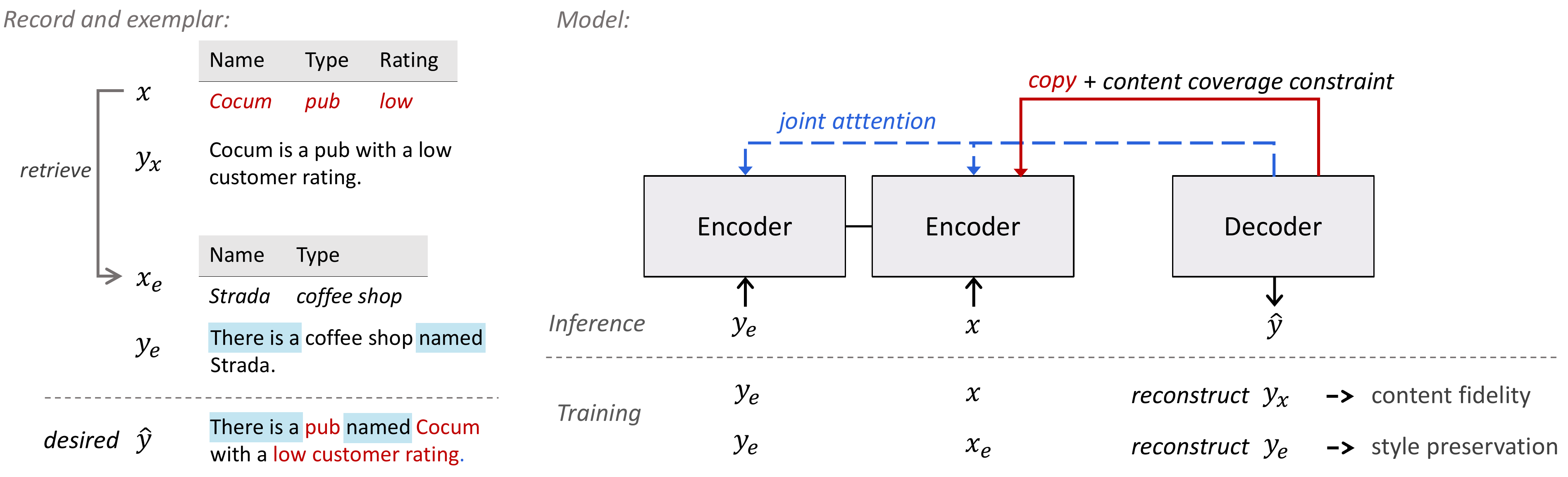}
%   \vspace{-20pt}
\caption{A (simplified) data example and retrieval {\bf (left)} and the model overview {\bf (right)}. 
The proposed approach uses a hybrid attention-copy mechanism, and is learned with weak supervisions and a content coverage constraint.
}
% \vspace{-0.4cm}
\label{fig:model}
\end{figure*}

\section{The Approach}\label{sec:model}

Denote the proposed neural model as $p_\theta(\bm{y}|\bm{x},\bm{y}_{e})$. The model has a hybrid attention-copy mechanism (sec~\ref{sec:model:hybrid}) for differentiated treatment of source content and style exemplar. We learn the model by constructing weak supervisions from the available non-parallel data (sec~\ref{sec:model:weak}), and further encourage accurate content description with a content coverage constraint (sec~\ref{sec:model:constraint}). Figure~\ref{fig:model} presents an overview of the approach.

\subsection{Hybrid Attention-Copy Architecture}\label{sec:model:hybrid}

%As shown in Figure~\ref{fig:model}, 
The overall architecture of the neural model consists of two encoders and one decoder. The two encoders extract the representation of the data record $\bm{x}$ and exemplar $\bm{y}_{e}$, respectively. Concretely, for each field in $\bm{x}$, we concatenate the embedding vectors of the field and its value, and feed the sequence of field-value embeddings to the encoder.
%for each data tuple in a record, we first concatenate the embedding vectors of all fields in the tuple, and feed the combined embedding to the encoder. 
% There is no need to specify a particular order of the tuples as their fields have specified the associated player or team.

The decoder generates the output sentence with a hybrid attention-copy mechanism. In particular, the decoder applies joint attention over both $\bm{y}_{e}$ and $\bm{x}$, and uses a copy mechanism~\citep{gu2016incorporating} \emph{only} on the field values in the record $\bm{x}$. More concretely, at each step $t$, the decoder first attends jointly to the hidden states of both encoders, and obtains a decoding hidden state $\bm{h}_t$. The final output distribution is the weighted-sum of two distributions:
\begin{equation}
\small
\bm{P}^{(t)}_{out} = g_t \cdot \bm{P}^{(t)}_{V} + (1 - g_t) \cdot \bm{P}^{(t)}_{\bm{x}}
\label{eq:p-out}
\end{equation}
where $g_t$ is the probability of generating a token from the vocabulary; $\bm{P}^{(t)}_{V}$ is the generation distribution over the whole vocabulary; $\bm{P}^{(t)}_{\bm{x}}$ is the copy distribution over the field values in the record.
% All the quantities are computed based on the post-attention hidden state $\bm{h}_t$. 

%\subsection{Competing Learning Objectives}
\subsection{Learning with Weak Supervisions}\label{sec:model:weak}

The two problem goals, namely content fidelity and style embodiment, are complementary and to some extent competitive. We derive weak forms of supervisions for each of them, respectively, based on the corpus of record-description pairs available.

\paragraph{Exemplar Retrieval}
% First, for each record $\bm{x}$, we automatically
% construct the exemplar $\bm{y}_{e}$ through retrieval. Specifically, we use $\bm{x}$ to retrieve another record $\bm{x}_e$ based on their \emph{distance}, and use the description associated with $\bm{x}_e$ as the exemplar sentence $\bm{y}_{e}$ in training. The distance between two records is defined as the total number of different fields included in the records (i.e., fields that occur in one record but not the other). 
% We present more details of the retrieval in the Appendix~\ref{subsec:retireval}. 
% Figure~\ref{fig:model} gives an illustration of retrieved exemplar (with distance $=1$). We study the effect of training with exemplars of varying distances in the experiments. 

First, for each record $\bm{x}$, we automatically
construct the exemplar $\bm{y}_{e}$ through retrieval. Specifically, we use $\bm{x}$ to retrieve another record $\bm{x}_e$ based on their \emph{distance}, and use the description associated with $\bm{x}_e$ as the exemplar sentence $\bm{y}_{e}$ in training. 
We define the distance between $\bm{y}$ and $\bm{y}_{e}$ as follows:
\begin{equation}
\small
\mathcal D(\bm{y}, \bm{y}_{e})=\#[\bm{\mathcal{T}(\bm{x}) \cup \mathcal{T}(\bm{x}_{e})}] - \#[\bm{\mathcal{T}(\bm{x}) \cap \mathcal{T}(\bm{x}_{e})}].
\label{eq:retrieval}
\end{equation}
where $\mathcal{T}(\cdot)$ is the set of all fields in the record; $\#[\cdot]$ represents the number of fields in the set.
% The distance between two records is defined as the total number of different fields included in the records (i.e., fields that occur in one record but not the other). 
% We present more details of the retrieval in the Appendix~\ref{subsec:retireval}. 
Figure~\ref{fig:model} gives an illustration of retrieved exemplar (with distance $=1$). We study the effect of training with exemplars of varying distances in the experiments.

%We note that one could also plug in other retrieval methods such as using a neural ranker~\cite{cao-etal-2018-neural}. 
%More concretely, according to the definition in sec~\ref{sec:task}, we randomly select a record that has overlapping fields with $\bm{x}$. The corresponding description of the selected record serves as $\bm{y}_{sty}$ in training. Let $\bm{x}_{sty}$ denote the selected record. We present more details of the retrieval in the appendix.

\paragraph{Content Objective}
Given the retrieved results, we next tackle content fidelity. Consider the description associated with $\bm{x}$, which, though not following the desired style of $\bm{y}_{e}$, has accurately presented the content in $\bm{x}$. Denote the description as $\bm{y}_{x}$. We thus devise
the first learning objective that reconstructs $\bm{y}_{x}$ given $(\bm{x}, \bm{y}_{e})$, in order to provide the model with the hints on how the $\bm{x}$ content can be presented in natural language: 
%As defined in Section~\ref{sec:task}, the task has two simultaneous goals, namely data fidelity and style preservation. The two goals are in a sense competitive with each other. We base our unsupervised learning on this competitive relation.
%
% We make use of the side information $(\bm{y}_{aux}, \bm{x}_{sty})$ during training. Concretely, as the auxiliary sentence $\bm{y}_{aux}$ was originally written by humans to describe the content $\bm{x}$ and thus can be seen to have the maximum data fidelity, we devise the first objective that reconstructs $\bm{y}_{aux}$ given $(\bm{x}, \bm{y}_{sty})$:
\begin{equation}
\small
\mathcal{L}_{content}(\bm{\theta}) = \log p_\theta( \bm{y}_{x} | \bm{x}, \bm{y}_{e} ).
\label{eq:obj-content}
\end{equation}
%where $f_\theta$ is the proposed neural model. We call it the content fidelity sub-objective.

\paragraph{Style Objective}
For the second goal of style embodiment, we want to encourage the model to generate sentences in a similar form of $\bm{y}_{e}$. To this end, we notice that, if we feed the model with the exemplar sentence $\bm{y}_{e}$ and its corresponding record $\bm{x}_{e}$, then by definition the desired output would be $\bm{y}_{e}$ itself. 
%% (as $\bm{y}_{sty}$ describes content $\bm{x}_{sty}$, and is in the same style of itself). 
We thus devise the second learning objective that reconstructs $\bm{y}_{e}$ given $(\bm{x}_{e}, \bm{y}_{e})$:
\begin{equation}
\small
\mathcal{L}_{style}(\bm{\theta}) = \log p_\theta( \bm{y}_{e} | \bm{x}_{e}, \bm{y}_{e} ).
\label{eq:obj-style}
\end{equation}
%We call it the style preservation sub-objective. 
The objective essentially treats the exemplar sentence encoder and the decoder together as an auto-encoding module, which effectively drives the decoder to reproduce the exemplar's characteristics. 

%We have developed the two learning objectives for the two task goals, respectively. 
\paragraph{Joint Training}
The above two learning objectives are competitive with each other such that, by combining them and optimizing jointly, the model is encouraged to learn to balance between content fidelity and style embodiment. A similar learning strategy of dividing a learning problem into multiple competitive objectives has also been used in previous work such as text style transfer~\citep{hu2017controllable,shen2017style}. 
%and controlled chatbots~\citep{tang2019target}.
More formally, the above two objectives are coupled together to train the model as follows:
\begin{equation}
\small
\mathcal{L}_{joint}(\bm{\theta}) = \lambda \mathcal{L}_{content}(\bm{\theta}) + (1-\lambda) \mathcal{L}_{style}(\bm{\theta}),
\label{eq:obj-joint}
\end{equation}
where $\lambda\in(0,1)$ is the balancing weight.

\subsection{Content Coverage Constraint}\label{sec:model:constraint}

As shown in the empirical study (section~\ref{sec:exp}), the above learning performs well in general yet sometimes still fall short of expressing the record accurately. We thus devise an additional learning constraint to enhance content fidelity.
% based on the intuition that each data tuple should usually be conveyed \emph{exactly once} in the generated sentence.
The intuition is that, given the copy mechanism over the record $\bm{x}$, each field value in $\bm{x}$ should be copied exactly once.
%Intuitively, the copy mechanism over the record $\bm{x}$ supports a simple yet effective way, that is, each tuple should be copied only once.
We thus minimize the following L2 constraint that encourages the temporally aggregated copy probability of each field value in $\bm{x}$ to be $1$:
\begin{equation}
\small
\mathcal{C}(\bm{\theta}) = \left\| \sum\nolimits_t \bm{P}_{\bm{x}}^{(t)} - \bm{1} \right\|^2
\label{eq:obj-coverage}
\end{equation}
where $\bm{P}_{\bm{x}}^{(t)}$, as defined in Eq.\eqref{eq:p-out}, denotes the copy distribution over all field values at decoding step $t$; and $\bm{1}$ is a vector with all ones. 
% It is still possible that tokens of the content values ``leak'' from the generation distribution $\bm{p}_{V}^{(t)}$ in Eq \eqref{eq:p-out}. We disable the leakage by masking out relevant words (particularly numbers) for each instance from the vocabulary.

% We note that prior work in other context has also explored the idea of \emph{coverage} through either architecture augmentation~\citep{tu2016modeling} or inference penalty~\citep{wu2016google}. We tried these techniques but did not obtain noticeable improvement. As shown in the experiments, the proposed explicit coverage constraint over  copy probability leads to significant performance gains.

The full model training objective with the constraint is thus written as:
\begin{equation}
\small
\mathcal{L}(\bm{\theta}) = \mathcal{L}_{joint}(\bm{\theta}) - \eta \cdot \mathcal{C}(\bm{\theta})
\label{eq:obj-full}
\end{equation}
where $\eta\geq0$ is the weight of constraint.

\begin{table}[t]
\small
\centering
\begin{tabular}
%{@{}r | L{14.3cm}@{}}
{@{}r|L{0.5cm} L{0.5cm} L{0.5cm}|L{0.5cm} L{0.5cm} l@{}}
\cmidrule[\heavyrulewidth]{1-7}
&\multicolumn{3}{c|}{\textbf{\scriptsize{Restaurant Recommend.}}
} & \multicolumn{3}{c}{\textbf{\scriptsize{NBA Reports}}} \\

& {\bf \scriptsize{Train}} & {\bf \scriptsize{Dev}} & {\bf \scriptsize{Test}} & {\bf \scriptsize{Train}} & {\bf \scriptsize{Dev}} & {\bf \scriptsize{Test}} \\ \cmidrule{1-7}
\scriptsize{\#Instances} &  \scriptsize{29,486} & \scriptsize{6,299} & \scriptsize{6,273}& \scriptsize{31,444} & \scriptsize{6,765} & \scriptsize{6,930} \\
%\#Tokens & 541452. 1644560 & 353477 & 363799 \\
\scriptsize{\#Tokens} &  \scriptsize{0.54M} & \scriptsize{0.12M} & \scriptsize{0.12M} & \scriptsize{7.88M} & \scriptsize{1.69M} & \scriptsize{1.75M} \\
\scriptsize{Avg Text Length} &  \scriptsize{18.36} & \scriptsize{18.34} & \scriptsize{18.35} & \scriptsize{25.07} & \scriptsize{25.10} & \scriptsize{25.32} \\
\scriptsize{\#Unique Fields}  & \scriptsize{8} & \scriptsize{8} & \scriptsize{8} & \scriptsize{34} & \scriptsize{34} & \scriptsize{34}\\
\scriptsize{Avg \#Fields}  & \scriptsize{5.38} & \scriptsize{5.38} & \scriptsize{5.35} & \scriptsize{4.32} & \scriptsize{4.31} & \scriptsize{4.35}\\
\cmidrule[\heavyrulewidth]{1-7}
\end{tabular}
\vspace{-10pt}
\caption{Statistics of the two datasets.}
% \vspace{-0.4cm}
\label{tab:data}
\end{table}

\section{Experiments}\label{sec:exp}

We study on two datasets in the restaurant recommendations and NBA reports domains, respectively. We conduct both automatic and human evaluations to assess model performance. Experiment results validate the proposed approach in learning an effective, balanced control of content and style.
%For automatic evaluation, we use two metrics to measure content fidelity and style preservation respectively. {\bf Experimental results show our approach controls the trade-off between two goals better than a variety of comparison methods.} 

\subsection{Datasets}\label{sec:dataset}
We derived and processed the two existing popular corpora as below. As defined in section~\ref{sec:task}, each resulting dataset contains record-description pairs. 
% We provide more details of data processing in Appendix~\ref{sec:datacreation}. 
Table~\ref{tab:data} shows the data statistics. 

%We further introduce two large datasets as the testbeds for the task, which are derived automatically from existing popular corpora. As in sec~\ref{sec:task}, each dataset contains a collection of record-description pairs. For evaluation purpose, in the test sets of both data, we accompany each record with a style reference sentence (and its record) as retrieved in a similar way described in sec~\ref{sec:model:weak}. We provide more details of data creation in the appendix. Table~\ref{tab:data} shows the data statistics.

%We now present two datasets developed for the task, whose statistics are summarized in Table~\ref{tab:data}.
%% The corpus is originally used for studying supervised game report generation which has attracted increasing research interest~\citep{nie2018operations,wiseman2017challenges}.
%{\bf The main two pre-processing steps are (1) extracting the associated records from the table, and (2) retrieving the style sentence. We also note that other data-to-text datasets, such as the WikiBio ~\citep{lebret-etal-2016-neural} and WebNLG~\citep{gardent-etal-2017-creating} dataset, can be adapted to the task as well via the above pre-processing steps.}

\paragraph{Restaurant Recommendations}  
The dataset is extracted from the E2E NLG challenge~\citep{dusek2019e2e}. A restaurant record can contain a subset of 8 fields, such as \emph{Eat Type}, \emph{Price Range}, and others. See Figure~\ref{fig:task} for an example record and the different possible ways of description.
%A recommendation sentence expressing the record content as 18 tokens on average.

% The restaurant dataset consists of the meaning representation (MR) and text description provided by the E2E NLG challenge~\citep{dusek2019e2e}. Following the NBA dataset, the MR is pre-processed to various data tuples, e.g., (\emph{eatType, coffee shop, Bibimbap\ House}). Note that the number of data tuples of each record is limited. Thus to make this dataset more challenging, the retrieved $\bm{x}_{sty}$  would have more different content types from $\bm{x}$ than the NBA dataset.  We provide more details about the data creation of both datasets in Appendix. 

\paragraph{NBA Reports} 
We extract the dataset from the NBA game corpus developed in~\citep{wiseman2017challenges}. The original corpus consists of box-score tables of NBA matches and the corresponding full-length match reports. 
%This dataset is derived from a recent large NBA game corpus~\citep{wiseman2017challenges} which consists of box-score tables and associated document-level report.
We first split each report into individual sentences and extract the associated information from the box-score table as the data record. The data contains 34 unique fields, such as \emph{Points}, \emph{Rebounds} , \emph{Field-Goal Percentage}, etc. Though the recorded fields look regular, the natural language descriptions are rich with variation. For example, for a field value \emph{Points: 14}, one could say \emph{``contributed 18 points''}, \emph{``reached double figures''}, or, fusing with other fields, \emph{``scored an amazingly efficient 18 points on 7-of-8 shooting''}, etc.

%As discussed in Section \ref{sec:task}, the content record contains various tuples. The resulting record-sentence pair is treated as a pair of $(\bm{x}, \bm{y}_{aux})$. The next step is to find a suitable style sentence $\bm{y}_{sty}$ for each content record $\bm{x}$. We retrieve another record-sentence pair from the training set, where the record retrieved would contain slightly different structures from $\bm{x}$. The retrieved record-sentence pair plays the role of $(\bm{x}_{sty}, \bm{y}_{sty})$ and is merged with $(\bm{x}, \bm{y}_{aux})$ to form an instance.

% Note that the common data-to-text dataset can be devised 

\begin{table*}[t]
\small
\centering
\begin{tabular}{@{}c r|l l l|l l l @{}} 
\cmidrule[\heavyrulewidth]{1-8}
\multicolumn{2}{c}{} &
\multicolumn{3}{c}{\textbf{Restaurant Recommendations}} & \multicolumn{3}{c}{\textbf{NBA Reports}}\\
\cmidrule[\heavyrulewidth]{1-8}
&  & \multicolumn{2}{c}{Content}  & Style &\multicolumn{2}{c}{Content}  & Style \\ %\cmidrule{1-4}
& {\bf Method} & {\bf \%Incl.-new} & {\bf \%Excl.-old} & {\bf m-BLEU} &  {\bf Precision} & {\bf Recall} & {\bf m-BLEU}\\ \cmidrule[\heavyrulewidth]{1-8}
\multirow{2}{*}{\bf Reference} &  AttnCopy-S2S &  78.88\tiny{$\pm$2.08}  & 99.71\tiny{$\pm$0.06} & 13.95\tiny{$\pm$0.52 } & 81.62\tiny{$\pm$3.25}  & 75.65\tiny{$\pm$7.42} & 45.5\tiny{$\pm$0.71}   \\ 
& Slot-filling & 61.23 & 66.2 & 100 & 56.69 & 71.34 & 100\\ 
%\cmidrule{1-4}
\cmidrule{1-8}% morecmidrules\cmidrule{1-8}

\multirow{2}{*}{\bf Baselines} & MAST  & 36.28\tiny{$\pm$0.25} & 37.06\tiny{$\pm$0.16} & {\bf 91.76\tiny{$\pm$0.28 }} & 23.06\tiny{$\pm$3.90} & 27.37\tiny{$\pm$3.88} & {\bf 95.43\tiny{$\pm$2.71}}\\
& AdvST & 51.64\tiny{$\pm$4.45} & 57.06\tiny{$\pm$4.44} & 76.02\tiny{$\pm$5.27 } & 67.37\tiny{$\pm$0.66} & 66.79\tiny{$\pm$1.43} & 64.67\tiny{$\pm$4.81} \\

\cmidrule{1-8} 
% \multicolumn{6}{l}{\quad\quad{\normalize \bf{Our results:}}} \\

\multirow{4}{*}{\bf Ours} 
& Transformer w/o Coverage & 60.03\tiny{$\pm$2.16} & 74.65\tiny{$\pm$2.69} & 77.81\tiny{$\pm$3.83 } & 62.58\tiny{$\pm$2.88} & 70.22\tiny{$\pm$3.58} & 81.75\tiny{$\pm$2.32 }\\
% + Copy & 65.76\tiny{$\pm$2.45} & 73.61\tiny{$\pm$0.08} & 81.10\tiny{$\pm$2.87}& 61.96\tiny{$\pm$0.23} & 25.31\tiny{$\pm$1.17} & 79.15\tiny{$\pm$1.49 } \\
& + Coverage &61.84\tiny{$\pm$1.31} &  81.14\tiny{$\pm$2.73} & 80.29\tiny{$\pm$0.35 } &  67.74\tiny{$\pm$0.79} &  {\bf 74.35\tiny{$\pm$1.22}} &  81.97\tiny{$\pm$2.87}\\\cmidrule{2-8}
& LSTM w/o Coverage & 60.83\tiny{$\pm$1.29} & 81.45\tiny{$\pm$1.10} & 78.91\tiny{$\pm$1.05 } & 68.74\tiny{$\pm$3.07} & 69.35\tiny{$\pm$3.30} & 79.88\tiny{$\pm$2.44}\\
& + Coverage & {\bf 65.02\tiny{$\pm$4.16}} & {\bf 82.53\tiny{$\pm$0.70}} & 82.92\tiny{$\pm$3.18 } &  {\bf 69.54\tiny{$\pm$1.16}} &  73.27\tiny{$\pm$1.18} & 80.66\tiny{$\pm$1.89} \\
\cmidrule[\heavyrulewidth]{1-8}
\end{tabular}
\vspace{-10pt}
\caption{Results of automatic evaluation, averaged over 3 runs $\pm$ one standard deviation.
The distance between the record and the exemplar is set to $\leq 5$ for exemplar retrieval (see the text). Methods in the first block are two reference approaches (Section~\ref{subsec:exper-setup}), i.e., \texttt{AttnCopy-S2S} for content fidelity and \texttt{Slot-filling} for style embodiment. For our method, we evaluate the variants with and without the coverage constraint (Section~\ref{sec:model:constraint}). 
The table highlights the best results in the blocks of Baselines and Ours under different metrics.} 
% Our two models with LSTM or Transformer achieve significant higher content fidelity and lower residue respectively compared to both rule-based and style transfer methods, and reach a high BLEU score in style preservation.}
\label{tab:auto-results}
%\vspace{-0.3cm}
\end{table*}
\begin{figure*}[t]
%\vspace{-10pt}
\centering
% \resizebox{6.8in}{3.8in}{}
\includegraphics[width=\linewidth]{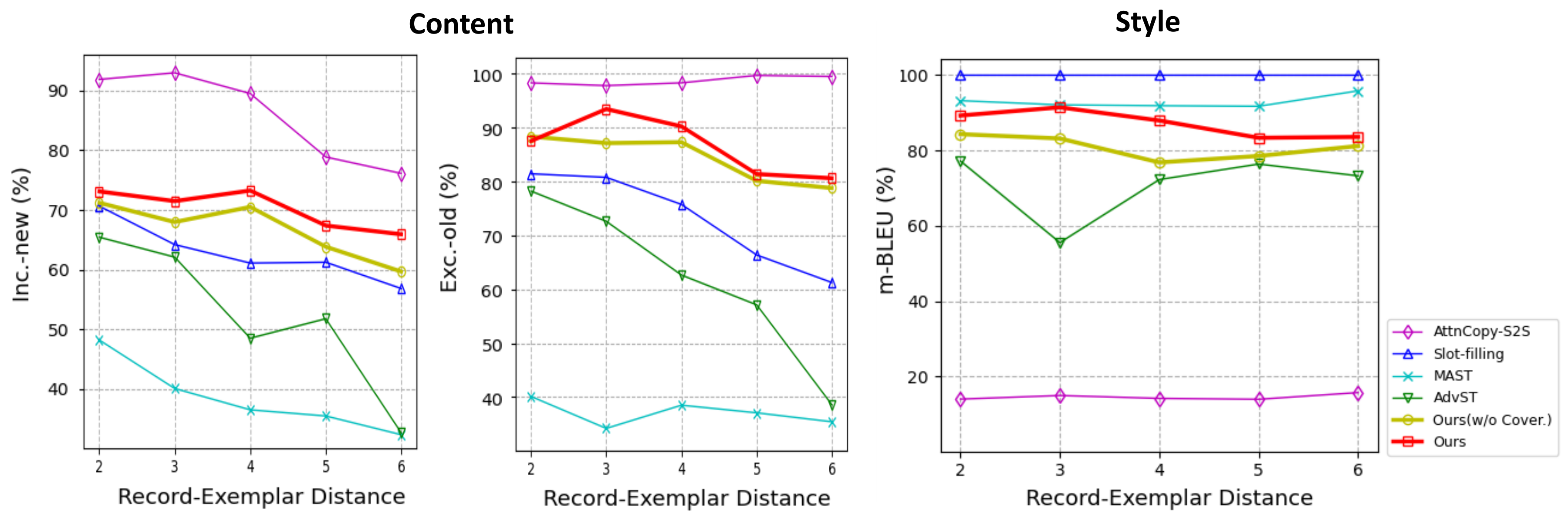}
\vspace{-15pt}
\caption{Effect of record-exemplar distance on model performance on the restaurant dataset. {\bf Left}: Content fidelity performance, including ``\%Inc-new'' and ``\%Exc-old''. {\bf Right}: Style embodiment performance by ``m-BLEU''.}
% \vspace{-0.4cm}
\label{fig:results}
\end{figure*}

\subsection{Setup}\label{subsec:exper-setup}

\paragraph{Comparison Approaches}\ \\
We compare with diverse approaches for a comprehensive analysis of the task and proposed approach:
\begin{itemize}[leftmargin=*]
    \item {\bf Reference for Content Fidelity: AttnCopy-S2S. }
    We first consider a conventional data-to-text model designed for only expressing the content. As style imitation is omitted, the method is expected to excel on content fidelity but fail on style control. 
    Specifically, we use a sequence-to-sequence model~\citep{sutskever2014sequence} augmented with the proposed attention-copy mechanism (Section~\ref{sec:model:hybrid}), which is trained supervisedly on the record-description pairs.
    %The method serves as a reference of the difficulty in expressing desired content.

    \item {\bf Reference for Style Embodiment: Slot-filling. }The second approach serving as a reference is a traditional slot-filling method that first removes the content words in the exemplar sentence $\bm{y}_{e}$ to make a template, and fills in the slots with respective values in the record $\bm{x}$. 
    As all content-independent tokens in $\bm{y}_{e}$ are preserved, the method is expected to perform well on style embodiment, but fail on content fidelity due to the possible mismatch between the exemplar sentences and desired content $\bm{x}$. We manually crafted a large set of slot-filling rules for each of the two datasets respectively.
    % (see Appendix~\ref{subsec:slotfilling} for more details). 

    % {\bf \item {\bf Rule-based Method. }A straightforward way for text content manipulation is to align $\bm{x}$ with $\bm{x}_{sty}$ and replace portions of $\bm{y}_{sty}$ with $\bm{x}$ in the style sentence. 
    % % More concretely, we first align $\bm{x}$ and $\bm{x}_{sty}$ through data types and then align $\bm{x}_{sty}$ and $\bm{y}_{sty}$ through data values, types and indicative tokens. The two alignments connect $\bm{x}$ and $\bm{y}_{sty}$, enabling us to swap appropriate text in $\bm{y}_{sty}$ to express content $\bm{x}$. 
    % Intuitively, rule-based method would perform well in preserving the style, as it merely replaces content related tokens without modifying other parts of the style sentence. However, this approach tends to leave out some new content and retain irrelevant information due to the different structures between $\bm{x}$ and $\bm{x}_{sty}$.} 
     
    \item {\bf Multi-Attribute Style Transfer (MAST)}~\citep{subramanian2019multiple}. We compare with a recent style transfer approach capable of manipulating multiple attributes. To apply to our task, we treat the field values in record $\bm{x}$ as separate attributes. The method is based on back-translation~\citep{sennrich2015improving} that first generates a target sentence $\bm{\hat{y}}$ conditioning on $(\bm{x}, \bm{y}_{e})$, and then treat it as the reference to reconstruct $\bm{y}_{e}$ conditioning on $(\bm{x}_{e}, \bm{\hat{y}})$. Auxiliary sentence $\bm{y}_{x}$ is used in an extra auto-encoding loss. 
    % The method is based on back-translation, as detailed in Appendix~\ref{subsec:MAST}.
    
    %The method is based on back-translation~\citep{sennrich2015improving} that first generates a target sentence $\bm{\hat{y}}$ conditioning on $(\bm{x}, \bm{y}_{sty})$, and then treat it as the reference to reconstruct $\bm{y}_{sty}$ conditioning on $(\bm{x}_{sty}, \bm{\hat{y}})$. Auxiliary sentence $\bm{y}_{aux}$ is used in an extra auto-encoding loss.
    
    \item {\bf Adversarial Style Transfer (AdvST)}~\citep{logeswaran2018content}. As another style transfer approach for multiple attributes, the model incorporates back-translation with adversarial training to disentangle content and style representations.
    
    %\item {\bf Ours w/o Coverage. } For ablation study, we compare with a variant of our approach that omits the content coverage constraint (sec~\ref{sec:model:constraint}). 
    %That is, the model is trained by maximizing only Eq.\eqref{eq:obj-joint}.
    
    % \item {\bf Ours Transformer w/o Copy. } We empirically found that disenabling the self-attention to $\bm{y}_{sty}$ only at the final decoding block can achieve the similiar copying effects. The results are reasonable , since this operation enables it attending to $\bm{x}$ to the greatest extent. Hence, we compare with this kind of standard transformer model.
    
    % \item {\bf Ours Transformer w/o Coverage. }For ablation study, we compare with our transformer model variant that omits the content coverage constraint . That is, the model is trained by maximizing only Eq \eqref{eq:obj-joint}.
\end{itemize}

\begin{table*}[t]
\centering
\small
\begin{tabular}{@{}r|l l l|l l l@{}}
\cmidrule[\heavyrulewidth]{1-7}
\multicolumn{1}{c}{}&\multicolumn{3}{c}{\textbf{Restaurant Recommendations}} &\multicolumn{3}{c}{\textbf{NBA Reports}} \\
\cmidrule{1-7}
 & \multirow{2}{1.2cm}{{\bf Content Fidelity}}  & \multirow{2}{1.2cm}{{\bf Style Embody}} & & \multirow{2}{1.2cm}{{\bf Content Fidelity}}  & \multirow{2}{1.2cm}{{\bf Style Embody}}\\
{\bf Model} & & &{\bf Fluency} & & &{\bf Fluency} \\
\cmidrule{1-7}
Slot-filling       & 3.36  & {\bf 5.00}  & {\bf 4.70} & 2.79  & {\bf 5.00}  & {\bf 4.86} \\
AdvST    & 3.56  & 4.24  & 4.02 & 2.88  & 4.00  & 4.09  \\
\cmidrule{1-7}
% \multicolumn{6}{l}{\quad\quad\quad\quad{\normalize \bf{Ours:}}}\\
Ours, LSTM w/o Coverage   & 3.91  & 4.38  & 4.58 & 3.43  & 4.13  & 4.59\\
Ours, LSTM        & {\bf 4.28}  & 4.73  & 4.54 & {\bf 3.88}  & 4.53  & 4.52\\
\cmidrule[\heavyrulewidth]{1-7}
& {\bf Ours Better}&{\bf No Prefer.}&{\bf Ours Worse} &{\bf Ours Better} & {\bf No Prefer.} &{\bf Ours Worse}
\\\cmidrule{1-7}
Slot-filling & {\bf 64.1\%} & 18.6\% & 17.3\% & {\bf 67.5\%} & 17.5\% & 15.0\%\\
AdvST  & {\bf 70.4\%} & 14.3\% & 15.2\% & {\bf 68.8\%} & 17.5\% & 13.8\% \\
Ours, LSTM w/o Coverage & {\bf 52.0\%} & 26.7\% & 21.3\% & {\bf 51.3\%} & 32.5\% & 16.3\%\\
\cmidrule[\heavyrulewidth]{1-7}
\end{tabular}
\vspace{-10pt}
\caption{Results of human evaluation. 
Each metric achieves an average Pearson correlation coefficient $\ge$0.73, showing a reasonable inter-annotator agreement. Our improvement in terms of mean annotator ratings is statistically significant (p\textless0.01, t-test).
{\bf Top:} Scoring three aspects on a 5-point Likert scale. 
{\bf Bottom:} Ranking the generations from  pairs of models. We use our LSTM-based full model to compare with other methods.} 
% Our model wins on more than 50\% instances compared to each of other models on two datasets.}
\vspace{-0.33cm}
\label{tab:human-results}
\end{table*}

\paragraph{Model Configurations}\ \\
We studied both LSTM~\cite{hochreiter1997long} and Transformer~\citep{vaswani2017attention} architectures. For LSTM, we use a single layer with the Luong attention~\citep{luong2015effective} and copy mechanism~\cite{gu2016incorporating}. For Transformer, use the recent copy-augmented variant following~\citep{su-etal-2019-improving} with 3 blocks.
%we use 3-blocks augmented with the pointer network following~\citep{su-etal-2019-improving}. 
%Both the embedding dimension and hidden dimension are set to 384. 
During training, we first set $(\lambda=0, \eta=0)$ to pre-train the model so that it captures the full characteristics of the exemplar sentence. We then switch to $(\lambda=0.2, \eta=1.0)$ for full training. Adam optimization~\citep{kingma2014adam} is used with an initial learning rate of 0.001.
%and gradient norm clipping of 15. 
At inference time, we use beam search with the width 5 and the maximum decoding length 50. 
%The maximum decoding length is set to 50. 

\subsection{Automatic Evaluation}\label{subsec:auto-eval}
%We first adopt automatic metrics to evaluate the model performance quantitatively. 

\paragraph{Metrics}\ \\
Automatic evaluation of the task is an open and challenging problem. We use several quantitative metrics for the two goals of the task, namely content fidelity and style embodiment. 
%The desired solution should achieve the trade-off between both goals well.
\begin{itemize}[leftmargin=*]
    \item {\bf Content fidelity.} 
    For the NBA data, we follow the original work~\citep{wiseman2017challenges} and use information extraction (IE) to measure content fidelity. Given a generated sentence $\bm{\hat{y}}$ and the input data record $\bm{x}$, we extract field values from $\bm{\hat{y}}$ with an IE tool and compute the precision and recall against $\bm{x}$. We use the IE model provided in~\citep{wiseman2017challenges}, which achieves 81\% precision and 86\% recall on the test set.
    
    We found IE on the restaurant data is too difficult to serve as a reliable metric, because the descriptions are less structured. We thus instead train a BERT-based binary classifier 
    % (see Appendix~\ref{sec:classifier} for more details) 
    to evaluate whether a field value is expressed in the generated sentence, which achieves 94\% classification accuracy on  the test set. We apply the classifier and compute both the percentage of desired $\bm{x}$ field values expressed in the generation (\emph{\%Incl.-new}) and the percentage of original content in $\bm{y}_{e}$ (or equivalently, $\bm{x}_{e}$) removed from the generation (\emph{\%Excl.-old}). The higher both numbers, the more faithfully the generation describes $\bm{x}$.
    
    % {\bf For the restaurant dataset, we adopt two metrics: content residue and fidelity. The \emph{residue} counts the proportion of old data tuples that retains in the generated sentences, while the \emph{fidelity} computes the new ones' that be expressed.
    % % to count quantities of new and old contents records conveyed by generated sentences respectively. 
    % More concretely, we train a Bert-based binary classifier in advance to determine whether the data tuple $\bm{x}_i$ is expressed in the generated sentence $\bm{\hat{y}}$. We create the training data via concatenating each $\bm{x}_i$ with $\bm{\hat{y}}$, e.g., \emph{customerRating: 1 star $|$ Cocum is a pub with a low rating.} The classifier achieves 94\% accuracy on the test set and thereby can be used reasonably for evaluation.} 
    % A desired model should express more new content records and less old ones. Hence, the lower residue represent the better performance of manipulating the old content. 
    
    \item {\bf Style embodiment. }
    %A generated sentence is desired to retain stylistic properties, such as word choice and expressions order of the input style sentence. 
    Imitating the exemplar style involves inheriting the sentence structure, word choices, and other surface forms of $\bm{y}_{e}$.
    Inspired by the text style transfer literature~\citep{subramanian2019multiple,yang2018unsupervised}, we measure the BLEU score between the generated and the exemplar sentences. To reduce the influence of the change of content tokens, we mask in both sentences all obvious content tokens, e.g., player/team names and numbers, by replacing them with a special token \texttt{<M>}. We denote the metric as \emph{m-BLEU}. This guarantees the reference approach, namely the slot-filling method, achieves an m-BLEU score of 100. 
    %{\bf We also empirically note that the BLEU score between $\bm{y}_{aux}$ and $\bm{\hat{y}}$ is too low for reference.}
    % In this way, the above rule-based method would have a maximum BLEU score of 100, which is consistent with our intuition above.
\end{itemize}

%\vspace{-20pt}
\begin{table*}[!t]
\centering
\begin{spacing}{0.7}
\small
\begin{tabular}{@{}r | l@{}}
\cmidrule[\heavyrulewidth]{1-2}
  {\bf Content Record} &  
 \begin{tabular}{@{}l l l l l l l l l @{}}
 {\bf Name} & {\bf EatType} & {\bf Food} & {\bf PriceRange} & {\bf CustomRating} & {\bf FamilyFriendly
} \\
 Cocum & coffee shop & Italian & \pounds20-25 & high & family friendly
 \end{tabular} 
 \\ 
\cmidrule[\heavyrulewidth]{1-2}
{\bf Exemplar 1} & \style{Looking for} French food \style{near} Zizzi? \style{Come try} Strada, \style{which has a} 3-star \style{customer rating and priced} lowly. \\[2pt] \cmidrule{1-2}
Slot filling & Looking for Italian \error{[...]} food \error{near Zizzi}? Come try \error{[...]} Cocum, which has a high customer rating and \\
& priced \pounds20-25.\\[2pt] \cmidrule{1-2}
AdvST &  For Italian \error{[...]} place \error{near Zizzi}? Come try \error{[...]} Cocum, which has a high customer rating \error{with} priced \\
& \pounds20-25. \\[2pt] \cmidrule{1-2}
Ours & \style{Looking for} an Italian coffee shop? \style{Come try} family-friendly Cocum, \style{which has a} high \style{customer rating}\\
& \style{and priced} \pounds20-25. \\[2pt]
\cmidrule[\heavyrulewidth]{1-2}

{\bf Exemplar 2} & \style{Along the} riverside \style{near} Cafe Rouge\style{, there is a } Japanese \style{food place called} The Golden Curry. \style{It has an }\\
& average \style{customer rating since it is} not \style{a} family-friendly \style{environment.}   \\[2pt] \cmidrule{1-2}
 Slot-filling & Along the \error{riverside} near \error{Cafe Rouge [...]}, there is \error{a} Italian food \error{[...]} place called Cocum. It has \error{an} high\\ & customer rating since it is \error{not} a family-friendly environment. \\[2pt] \cmidrule{1-2}
AdvST & Along the \error{riverside} near \error{the Ranch [...]}, there is \error{a} Italian food \error{[...]} place called Cocum. It has \error{[...]} high \\
& customer rating since it is \error{not} a family-friendly environment. \\[2pt] \cmidrule{1-2}
% Ours w/o Cover. & Along the \pounds20-25 near, there is \error{a} Italian food place called Cocum. It has \error{an} high customer rating. It is \error{not}\\
% & a family\_friendly environment.\\[2pt] \cmidrule{1-2}
Ours & Priced \pounds20-25\style{, there is} an Italian \style{food} coffee shop \style{called} Cocum. \style{It has} a high \style{customer rating since it is}\\
& \style{a} family-friendly \style{environment.}   \\[2pt]
\cmidrule[\heavyrulewidth]{1-2}

\cmidrule[\heavyrulewidth]{1-2}
{\bf Content Record}  &  
 \begin{tabular}{@{}l l l l l l l l l @{}}
 {\bf PLAYER} & {\bf PLAYER} & {\bf PLAYER} & {\bf PTS} \\
  Patrick & Dwight Howard & Harden & 10
 \end{tabular} 
 \\ \cmidrule[\heavyrulewidth]{1-2}
{\bf Exemplar} & \style{Both} J.J. Hickson \style{and} Timofey Mozgov \style{reached double - figures , scoring} 10 and 15 \style{points}. \\[2pt] \cmidrule{1-2}
Slot-filling & Both Patrick \error{[...]} and Dwight Howard reached double - figures , scoring 10 \error{and 15} points. \\[2pt] \cmidrule{1-2}
AdvST & Both \error{J.J. Hickson} \error{[...]} and Dwight Howard reached double - figures , scoring 10 \error{and 10} points.\\[2pt] \cmidrule{1-2}
% Ours w/o Cover. & Both Patrick, Dwight Howard and Harden reached double - figures , scoring 10 \error{and 10} points. \\[2pt] \cmidrule{1-2}
Ours &  
Patrick , Dwight Howard \style{and} Harden \style{reached double - figures , scoring} 10 \style{points.}\\[2pt]
\cmidrule[\heavyrulewidth]{1-2}
% {\bf Exemplar 2} & \style{Three} Phoenix \style{players reached double figures, with} Alex Len, Leandro Barbosa \style{and} Brandon Knight \\
% &\style{all amassed} 20 - plus \style{minutes} and 12 - plus \style{points in the contest.} \\[2pt] \cmidrule{1-2}
% Slot-filling & Three \error{Phoenix} players reached double figures, with Patrick, Dwight Howard and Harden all amassed \\
% & \error{20 - plus minutes} and 10 - plus points in the contest. \\[2pt] \cmidrule{1-2}
% AdvST & Three \error{Dwight Howard Phoenix} players reached double figures, with Patrick, Harden and \error{Brandon Knight} \\
% & all amassed \error{20 - plus minutes} and 10 - plus points in the contest.\\[2pt] \cmidrule{1-2}
% % Ours w/o Cover. & Three \error{Dwight Howard} players reached double figures, with \error{10}, Patrick and Harden all amassed minutes \\
% % & and 10 - plus points in the contest. \\[2pt] \cmidrule{1-2}
% Ours &  \style{Three players reached double figures, with } Patrick, Dwight Howard \style{and} Harden \style{all amassed} \error{minutes} \style{and}\\
% & 10 - plus \style{points in the contest.}\\[2pt]
% \cmidrule[\heavyrulewidth]{1-2}
\end{tabular}
% \vspace{-6pt}
\end{spacing}
\caption{Example outputs by different models given various exemplar sentences. Text of erroneous content and syntax are highlighted in \error{red}, where \error{[...]} indicates desired content that is missing. Text portions about the writing style in both exemplars and the generated sentences by our model are highlighted in \style{blue}.}
% \vspace{-0.2cm}
\label{tab:example-outputs}
\end{table*}

\paragraph{Study: Balance between Content and Style}\ \\
%We now compare the performance of different approaches in terms of the above metrics. 
Table~\ref{tab:auto-results} shows the automatic evaluation results on the two datasets. In this study, for exemplar retrieval (Section~\ref{sec:model:weak}), we set the distance between a record and an exemplar to be no larger than 5 both during training and when constructing test cases. That is, the record and the exemplar sentence can have 5 mismatched fields, which thus requires strong flexibility of the generation model to be able to automatically adapt the exemplar in order to describe the record accurately.

As expected, the reference methods excel only in one of the two aspects, respectively. Specifically, \texttt{AttnCopy-S2S} expresses the desired content well, yet is incapable of embodying the designated style (e.g., m-BLEU=13.95). On the contrary, the \texttt{Slot-filling} method achieves perfect style m-BLEU by definition, but falls short of adaptively described the desired content in an accurate way, as shown by the low content scores.
% The first block shows the two baseline models providing reference performance. 
%  \error{TODO: Here we need to clarify that both models only perform well in one goal.} Both models can only perform well in one goal. The AttnCopy-S2S model only concerns about content fidelity and achieves high content scores, while its BLEU score is particularly low. As discussed in Section ~\ref{subsec:exper-setup}, the rule-based approach can reach the maximum BLEU (100). However, its performance is unsatisfactory in terms of the content precision and residue. 
% 
The two style transfer approaches (\texttt{MAST} and \texttt{AdvST}) also fail in terms of content fidelity performance.
%fail the expectation, as their content fidelity performance is greatly inferior or merely comparable to the rule-based approach. 
This is partly because these models are built on a different task assumption (i.e., modifying independent textual attributes) and are incompetent in manipulating the structured content well.

Our proposed approach is able to better balance between content fidelity and style embodiment. For example, in terms of content fidelity, our approach with an LSTM architecture improves over the \texttt{Slot-filling} results by 16.3 on NBA content precision and 12.9 on Restaurant content \%Excl.-old. The approach meanwhile keeps a high style m-BLEU score of over 80. 
Regarding the ablation study, the results show the proposed content coverage constraint (Section~\ref{sec:model:constraint}) consistently improves both the content and style performance by a large margin. We note that the LSTM and transformer architectures perform comparably, with LSTM slightly better on the restaurant dataset. We speculate that the copy mechanism of LSTM~\cite{gu2016incorporating} is slightly more effective than that of transformer~\cite{su-etal-2019-improving}.

%Concretely, the proposed LSTM-based model obtains 16.3/12.9 points absolute improvement over the rule-based results on residue/precision, 
%% substantially improving over other approaches (e.g., with a 12-point precision boost in comparison with the rule-based baseline) except for AttnCopy-S2S which has failed in style control. while obtaining a high BLEU score of over 80. 
%The superior performance of the proposed model to the ablated one demonstrates the effectiveness of the content coverage constraint (Eq \ref{eq:obj-coverage}). By encouraging the model to mention each of the data tuples exactly once, the model achieves higher content fidelity with less style ``sacrificed''.
%{\bf We note that our proposed LSTM-based model is slightly superior to the Transformer-based one. The results may be owing to that the text length in both datasets is short, while transformer excels more at the process of long text.} 
%  Besides, the tens of thousands level amount of data may not support the huge parameters of transformer well.

\paragraph{Study: Effect of Record-Exemplar Distance}\ \\
We then study how well the different methods would perform when given exemplars of varying distances (mismatchness) to the records. Figure~\ref{fig:results}
 show the content and style results under different distances. We can see that, as the exemplars deviate more from the structure of the records, the model performance drops since it is getting harder to automatically adapt the exemplars to express the desired content. For example, the ``\%Excl.old'' score (middle panel) of the methods \texttt{Slot-filling} and \texttt{AdvST} decreases quickly. Our approach maintains a more stable performance and keeps a better content-style balance. The results also show the proposed content coverage constraint consistently offers enhanced performance.

\subsection{Human Evaluation}\label{sec:human-eval}
We also perform human evaluation for a more thorough and accurate comparison. 
%Each instance for evaluation contains the source input and outputs of different approaches. 
Following the experimental settings in prior work~\citep{subramanian2019multiple,logeswaran2018content,shen2017style}, we undertake two types of human evaluation: 
%(1) 120 instances are distributed evenly to 3 evaluators, who are required to score generated sentences in three aspects: content fidelity, style preservation and language fluency. Each score is from 1 (strongly bad) to 5 (strongly good); 
(1) We ask three human annotators to score generation results in three aspects, namely content fidelity, style embodiment, and sentence fluency, on a 5-point Likert scale. 
(2) We present to each annotator a pair of generated sentences, one from our model and the other from a comparison method, then ask the annotator to rank the two sentences by considering the above criteria jointly. Annotators can also choose ``no preference'' if the sentences are equally good or bad. For each study, we evaluated on 80 test instances. We use the LSTM architecture as it outperforms the transformer slightly in the automatic evaluation. We compare with the \texttt{Slot-filling} method, \texttt{AdvST} (which is better than \texttt{MAST} in automatic evaluation), and our variant without the coverage constraint.
%each evaluator is distributed 50 test instances and compare our model with three comparison approaches.
% the rule-based method, AdvST style transfer model (which has shown better performance than the other style transfer model MAST) and the model variant without coverage constraint.

Table~\ref{tab:human-results} shows the results. 
%For each metric, the average Pearson correlation coefficient of the scores given by three evaluators is greater than 0.73, which ensures the inter-annotator agreement. 
From the top block, as discussed above, the \texttt{Slot-filling} method performs well in terms of style embodiment and fluency. However, its content fidelity is extremely weak. In contrast, our model achieves a better balance across the three criteria, by obtaining the best performance on content fidelity and reasonably high scores on both style embodiment and fluency.
The fluency of our full model is slightly inferior to the variant without the coverage constraint, which is not unexpected since the full model modifies more portions of the exemplar sentences, which would result in minor language mistakes.

The bottom block of Table~\ref{tab:human-results} shows the human ranking results. We can see that our model consistently outperforms the comparison methods with over 50\% wins on both datasets.

\subsection{Qualitative Study}
% We take a closer look at the model performance by studying generated sentences from different models.
Table~\ref{tab:example-outputs} shows samples on two test cases. We can see that the proposed full model performs superior to other approaches in effectively retaining the desired style and describing the content. 
For example, in the first two examples, other approaches often fail to remove the redundant content (e.g., ``near Zizzi'' or ``riverside'') from the generation while neglecting desired fields in the record. The proposed model performs better by adaptively adding and deleting text portions for accurate content description.
%though sometimes it fails to ensure the grammatical correctness (e.g., the choice of the article). 
Similarly, in the third case, both \texttt{Slot-filling} and \texttt{AdvST} fail to convey the new field value ``Harden'' given the exemplar, and leave in irrelevant information given the second one due to the different record structures between $\bm{x}$ and $\bm{x}_{e}$. 
%The variant without coverage constraint both copied the record ``10 points'' twice given various writing styles. 
In contrast, our full model generates the desired sentence. 

\section{Conclusion}\label{sec:conclude}
We have studied the new problem of data-to-text generation with style imitation. We developed a new approach with an attention-copy mechanism, weakly supervised learning, and a content coverage constraint. Experiments show the approach achieves a good balance between content fidelity and style control, and is flexible to adapt exemplars that do not match the record perfectly. We are interested in applying the style imitation approach to control longer paragraphs given full data tables.

\balance
\bibliography{emnlp2020}

\begin{thebibliography}{40}
\expandafter\ifx\csname natexlab\endcsname\relax\def\natexlab#1{#1}\fi

\bibitem[{Angeli et~al.(2010)Angeli, Liang, and Klein}]{angeli2010simple}
Gabor Angeli, Percy Liang, and Dan Klein. 2010.
\newblock A simple domain-independent probabilistic approach to generation.
\newblock In \emph{EMNLP}, pages 502--512.

\bibitem[{Cao et~al.(2018)Cao, Li, Li, and Wei}]{cao2018retrieve}
Ziqiang Cao, Wenjie Li, Sujian Li, and Furu Wei. 2018.
\newblock Retrieve, rerank and rewrite: Soft template based neural
  summarization.
\newblock In \emph{ACL}, pages 152--161.

\bibitem[{Chen et~al.(2019)Chen, Tang, Wiseman, and
  Gimpel}]{chen2019controllable}
Mingda Chen, Qingming Tang, Sam Wiseman, and Kevin Gimpel. 2019.
\newblock Controllable paraphrase generation with a syntactic exemplar.
\newblock In \emph{ACL}.

\bibitem[{Deriu and Cieliebak(2018)}]{deriu-cieliebak-2018-syntactic}
Jan~Milan Deriu and Mark Cieliebak. 2018.
\newblock \href {https://doi.org/10.18653/v1/W18-6503} {Syntactic manipulation
  for generating more diverse and interesting texts}.
\newblock In \emph{INLG}, pages 22--34.

\bibitem[{Dou et~al.(2018)Dou, Qin, Wang, Yao, and Lin}]{dou2018data2text}
Longxu Dou, Guanghui Qin, Jinpeng Wang, Jin-Ge Yao, and Chin-Yew Lin. 2018.
\newblock Data2text studio: Automated text generation from structured data.
\newblock In \emph{EMNLP}, pages 13--18.

\bibitem[{Du{\v{s}}ek et~al.(2019)Du{\v{s}}ek, Novikova, and
  Rieser}]{dusek2019e2e}
Ond\v{r}ej Du{\v{s}}ek, Jekaterina Novikova, and Verena Rieser. 2019.
\newblock Evaluating the state-of-the-art of end-to-end natural language
  generation: {The} {E2E} {NLG} {Challenge}.
\newblock \emph{arXiv preprint arXiv:1901.11528}.

\bibitem[{Gehrmann et~al.(2018)Gehrmann, Dai, Elder, and
  Rush}]{gehrmann2018end}
Sebastian Gehrmann, Falcon~Z Dai, Henry Elder, and Alexander~M Rush. 2018.
\newblock End-to-end content and plan selection for data-to-text generation.
\newblock \emph{arXiv preprint arXiv:1810.04700}.

\bibitem[{Gu et~al.(2016)Gu, Lu, Li, and Li}]{gu2016incorporating}
Jiatao Gu, Zhengdong Lu, Hang Li, and Victor~OK Li. 2016.
\newblock Incorporating copying mechanism in sequence-to-sequence learning.
\newblock \emph{ACL}.

\bibitem[{Hashimoto et~al.(2018)Hashimoto, Guu, Oren, and
  Liang}]{hashimoto2018retrieve}
Tatsunori~B Hashimoto, Kelvin Guu, Yonatan Oren, and Percy~S Liang. 2018.
\newblock A retrieve-and-edit framework for predicting structured outputs.
\newblock In \emph{NeurIPS}, pages 10073--10083.

\bibitem[{Hochreiter and Schmidhuber(1997)}]{hochreiter1997long}
Sepp Hochreiter and J{\"u}rgen Schmidhuber. 1997.
\newblock Long short-term memory.
\newblock \emph{Neural computation}.

\bibitem[{Hu and Xing(2020)}]{hu2020learning}
Zhiting Hu and Eric~P Xing. 2020.
\newblock Learning from all types of experiences: A unifying machine learning
  perspective.
\newblock In \emph{KDD}.

\bibitem[{Hu et~al.(2017)Hu, Yang, Liang, Salakhutdinov, and
  Xing}]{hu2017controllable}
Zhiting Hu, Zichao Yang, Xiaodan Liang, Ruslan Salakhutdinov, and Eric~P Xing.
  2017.
\newblock Toward controlled generation of text.
\newblock In \emph{ICML}.

\bibitem[{Iso et~al.(2019)Iso, Uehara, Ishigaki, Noji, Aramaki, Kobayashi,
  Miyao, Okazaki, and Takamura}]{iso-etal-2019-learning}
Hayate Iso, Yui Uehara, Tatsuya Ishigaki, Hiroshi Noji, Eiji Aramaki, Ichiro
  Kobayashi, Yusuke Miyao, Naoaki Okazaki, and Hiroya Takamura. 2019.
\newblock \href {https://doi.org/10.18653/v1/P19-1202} {Learning to select,
  track, and generate for data-to-text}.
\newblock In \emph{ACL}, pages 2102--2113, Florence, Italy. Association for
  Computational Linguistics.

\bibitem[{Iyyer et~al.(2018)Iyyer, Wieting, Gimpel, and
  Zettlemoyer}]{iyyer2018adversarial}
Mohit Iyyer, John Wieting, Kevin Gimpel, and Luke Zettlemoyer. 2018.
\newblock Adversarial example generation with syntactically controlled
  paraphrase networks.
\newblock In \emph{NAACL}.

\bibitem[{Jagfeld et~al.(2018)Jagfeld, Jenne, and
  Vu}]{jagfeld-etal-2018-sequence}
Glorianna Jagfeld, Sabrina Jenne, and Ngoc~Thang Vu. 2018.
\newblock \href {https://doi.org/10.18653/v1/W18-6529} {Sequence-to-sequence
  models for data-to-text natural language generation: Word- vs.
  character-based processing and output diversity}.
\newblock In \emph{INLG}, pages 221--232.

\bibitem[{Kingma and Ba(2014)}]{kingma2014adam}
Diederik~P Kingma and Jimmy Ba. 2014.
\newblock Adam: A method for stochastic optimization.
\newblock \emph{arXiv preprint arXiv:1412.6980}.

\bibitem[{Kondadadi et~al.(2013)Kondadadi, Howald, and
  Schilder}]{kondadadi-etal-2013-statistical}
Ravi Kondadadi, Blake Howald, and Frank Schilder. 2013.
\newblock A statistical {NLG} framework for aggregated planning and
  realization.
\newblock In \emph{ACL}, pages 1406--1415.

\bibitem[{Kukich(1983)}]{kukich-1983-design}
Karen Kukich. 1983.
\newblock \href {https://doi.org/10.3115/981311.981340} {Design of a
  knowledge-based report generator}.
\newblock In \emph{ACL}, pages 145--150.

\bibitem[{Logeswaran et~al.(2018)Logeswaran, Lee, and
  Bengio}]{logeswaran2018content}
Lajanugen Logeswaran, Honglak Lee, and Samy Bengio. 2018.
\newblock Content preserving text generation with attribute controls.
\newblock In \emph{NeurIPS}, pages 5108--5118.

\bibitem[{Luong et~al.(2015)Luong, Pham, and Manning}]{luong2015effective}
Minh-Thang Luong, Hieu Pham, and Christopher~D Manning. 2015.
\newblock Effective approaches to attention-based neural machine translation.
\newblock \emph{arXiv preprint arXiv:1508.04025}.

\bibitem[{McRoy et~al.(2000)McRoy, Channarukul, and Ali}]{mcroy-etal-2000-yag}
Susan~W. McRoy, Songsak Channarukul, and Syed~S. Ali. 2000.
\newblock \href {https://doi.org/10.3115/1118253.1118293} {{YAG}: A
  template-based generator for real-time systems}.
\newblock In \emph{INLG}, pages 264--267.

\bibitem[{Pandey et~al.(2018)Pandey, Contractor, Kumar, and
  Joshi}]{pandey2018exemplar}
Gaurav Pandey, Danish Contractor, Vineet Kumar, and Sachindra Joshi. 2018.
\newblock Exemplar encoder-decoder for neural conversation generation.
\newblock In \emph{ACL}, pages 1329--1338.

\bibitem[{Peng et~al.(2019)Peng, Parikh, Faruqui, Dhingra, and
  Das}]{peng2019text}
Hao Peng, Ankur~P Parikh, Manaal Faruqui, Bhuwan Dhingra, and Dipanjan Das.
  2019.
\newblock Text generation with exemplar-based adaptive decoding.
\newblock In \emph{NAACL}.

\bibitem[{Power et~al.(2003)Power, Scott, and
  Bouayad-Agha}]{power2003generating}
Richard Power, Donia Scott, and Nadjet Bouayad-Agha. 2003.
\newblock Generating texts with style.
\newblock In \emph{International Conference on Intelligent Text Processing and
  Computational Linguistics}, pages 444--452. Springer.

\bibitem[{Puduppully et~al.(2019)Puduppully, Dong, and
  Lapata}]{puduppully-etal-2019-data}
Ratish Puduppully, Li~Dong, and Mirella Lapata. 2019.
\newblock \href {https://doi.org/10.18653/v1/P19-1195} {Data-to-text generation
  with entity modeling}.
\newblock In \emph{ACL}, pages 2023--2035.

\bibitem[{Reiter and Dale(1997)}]{reiter1997building}
Ehud Reiter and Robert Dale. 1997.
\newblock Building applied natural language generation systems.
\newblock \emph{Natural Language Engineering}, 3(1):57--87.

\bibitem[{Robin and McKeown(1996)}]{robin1996empirically}
Jacques Robin and Kathleen McKeown. 1996.
\newblock Empirically designing and evaluating a new revision-based model for
  summary generation.
\newblock \emph{Artificial Intelligence}, 85(1-2):135--179.

\bibitem[{Sennrich et~al.(2015)Sennrich, Haddow, and
  Birch}]{sennrich2015improving}
Rico Sennrich, Barry Haddow, and Alexandra Birch. 2015.
\newblock Improving neural machine translation models with monolingual data.
\newblock In \emph{ACL}.

\bibitem[{Shen et~al.(2017)Shen, Lei, Barzilay, and Jaakkola}]{shen2017style}
Tianxiao Shen, Tao Lei, Regina Barzilay, and Tommi Jaakkola. 2017.
\newblock Style transfer from non-parallel text by cross-alignment.
\newblock In \emph{NeurIPS}, pages 6830--6841.

\bibitem[{Su et~al.(2019)Su, Shen, Zhang, Sun, Hu, Niu, and
  Zhou}]{su-etal-2019-improving}
Hui Su, Xiaoyu Shen, Rongzhi Zhang, Fei Sun, Pengwei Hu, Cheng Niu, and Jie
  Zhou. 2019.
\newblock \href {https://doi.org/10.18653/v1/P19-1003} {Improving multi-turn
  dialogue modelling with utterance {R}e{W}riter}.
\newblock In \emph{Proceedings of the 57th Annual Meeting of the Association
  for Computational Linguistics}, pages 22--31, Florence, Italy. Association
  for Computational Linguistics.

\bibitem[{Subramanian et~al.(2019)Subramanian, Lample, Smith, Denoyer, Ranzato,
  and Boureau}]{subramanian2019multiple}
Sandeep Subramanian, Guillaume Lample, Eric~Michael Smith, Ludovic Denoyer,
  Marc'Aurelio Ranzato, and Y-Lan Boureau. 2019.
\newblock Multiple-attribute text rewriting.
\newblock In \emph{ICLR}.

\bibitem[{Sutskever et~al.(2014)Sutskever, Vinyals, and
  Le}]{sutskever2014sequence}
Ilya Sutskever, Oriol Vinyals, and Quoc~V Le. 2014.
\newblock Sequence to sequence learning with neural networks.
\newblock In \emph{NeurIPS}, pages 3104--3112.

\bibitem[{Tan et~al.(2020)Tan, Qin, Xing, and Hu}]{tan2020summarizing}
Bowen Tan, Lianhui Qin, Eric~P Xing, and Zhiting Hu. 2020.
\newblock Summarizing text on any aspects: A knowledge-informed
  weakly-supervised approach.
\newblock In \emph{EMNLP}.

\bibitem[{Tang et~al.(2019)Tang, Zhao, Xiong, Liang, Xing, and
  Hu}]{tang2019target}
Jianheng Tang, Tiancheng Zhao, Chengyan Xiong, Xiaodan Liang, Eric~P Xing, and
  Zhiting Hu. 2019.
\newblock Target-guided open-domain conversation.
\newblock In \emph{ACL}.

\bibitem[{Vaswani et~al.(2017)Vaswani, Shazeer, Parmar, Uszkoreit, Jones,
  Gomez, Kaiser, and Polosukhin}]{vaswani2017attention}
Ashish Vaswani, Noam Shazeer, Niki Parmar, Jakob Uszkoreit, Llion Jones,
  Aidan~N Gomez, {\L}ukasz Kaiser, and Illia Polosukhin. 2017.
\newblock Attention is all you need.
\newblock In \emph{NeurIPS}, pages 5998--6008.

\bibitem[{Weston et~al.(2018)Weston, Dinan, and Miller}]{weston2018retrieve}
Jason Weston, Emily Dinan, and Alexander~H Miller. 2018.
\newblock Retrieve and refine: Improved sequence generation models for
  dialogue.
\newblock \emph{arXiv preprint arXiv:1808.04776}.

\bibitem[{Wiseman et~al.(2017)Wiseman, Shieber, and
  Rush}]{wiseman2017challenges}
Sam Wiseman, Stuart~M Shieber, and Alexander~M Rush. 2017.
\newblock Challenges in data-to-document generation.
\newblock In \emph{EMNLP}.

\bibitem[{Wiseman et~al.(2018)Wiseman, Shieber, and Rush}]{wiseman2018learning}
Sam Wiseman, Stuart~M Shieber, and Alexander~M Rush. 2018.
\newblock Learning neural templates for text generation.
\newblock In \emph{EMNLP}.

\bibitem[{Yang et~al.(2018)Yang, Hu, Dyer, Xing, and
  Berg-Kirkpatrick}]{yang2018unsupervised}
Zichao Yang, Zhiting Hu, Chris Dyer, Eric Xing, and Taylor Berg-Kirkpatrick.
  2018.
\newblock Unsupervised text style transfer using language models as
  discriminators.
\newblock In \emph{NeurIPS}.

\bibitem[{Ye et~al.(2020)Ye, Shi, Zhou, Wei, and Li}]{ye2020variational}
Rong Ye, Wenxian Shi, Hao Zhou, Zhongyu Wei, and Lei Li. 2020.
\newblock Variational template machine for data-to-text generation.
\newblock In \emph{ICLR}.

\end{thebibliography}
\bibliographystyle{acl_natbib}

\clearpage

\end{document}

\maketitle

\appendix

% \section{For writing}
% \textbf{1. (human) evaluation that shows how much does this style-specification permits the variation of the output texts?}

% All of methods utilize the same reference sentence, so there may be no difference among our approach and baselines in term of variation/diversity of the output texts.

% \textbf{ 2. Difference with other E2E works? }

% The official E2E evaluation methods, e.g. BLEU and METEOR, mainly reflects the realization of the surface. For instance, with good surface realization but mediocre content fidelity may results in good BLEU, since most words in a sentence may be the stylistic word. However, our setting considers both factors more precisely and carefully.  

% \textbf{ 3. How much overlap is needed for the pair to be included?}

% Results in Table 2 and 3 both adopt the distance = 5. It's hard to say how much is the overlap since different pieces of record have different lengths of elements. But it's easy to compute the distance.

% \textbf{ 4. It seems like a slightly more complex template model could do better? }

% We've tried a more complex template model that conducts compulsory replacement. It achieves great content scores but poor fluency.

\section{Retrieval Approach}\label{subsec:retireval}

As described in Section 3 of the main text, the desirable exemplar sentence $\bm{y}_{e}$ would have the corresponding $\bm{x}_{e}$ that has analogous data types with $\bm{x}$.
Note that the ``style'' pair ($\bm{x}_{e}$, $\bm{y}_{e}$) is always retrieved from the training set.
Our approach uses $\bm{x}$ as the query data instance for retrieval.
We define the distance between $\bm{y}$ and $\bm{y}_{e}$ as follows:
\begin{equation}
\small
\mathcal D(\bm{y}, \bm{y}_{e})=\#[\bm{\mathcal{T}(\bm{x}) \cup \mathcal{T}(\bm{x}_{e})}] - \#[\bm{\mathcal{T}(\bm{x}) \cap \mathcal{T}(\bm{x}_{e})}].
\label{eq:retrieval}
\end{equation}
where $\mathcal{T}(\cdot)$ is the set of all data types in the record; $\#[\cdot]$ represents the number of elements in the set. Given a specific distance and $\bm{x}$, there may be many candidates $\bm{x}_{e}$ and to this end, we first select those with the same amount of the elements in the types set. Various distances reflect the difficult levels of the task. Then the corresponding sentence $\bm{y}_{e}$ to the retrieved $\bm{x}_{e}$ can be treated as an exemplar sentence.

\section{Dataset Preprocessing}\label{sec:datacreation}

\subsection{NBA Dataset}
Our NBA dataset is made from the \textsc{RotoWire} dataset ~\citep{wiseman2017challenges}, which consists of paired tables and paragraphs, each describing a NBA game. Each table is consisted of data records about box- and line-scores of basketball games, while each corresponding paragraph written by specialists is summarizing the game. In order to obtain desired $(\bm{x}, \bm{y})$ pairs, we first separate the paragraphs into sentences. Each sentence will be our $\bm{y}$. However, it is not so easy to obtain the corresponding $\bm{x}$ (i.e. data records): we have to find out which data records in the table of data records are described in $\bm{y}$.

To achieve this, we proposed a not precise but accurate enough rule based method to get $\bm{x}$ from $\bm{y}$. Two steps are performed in our method: Step 1, find out all entities (player, team, and city) and numbers in $\bm{y}$; Step 2, pick out candidate (entity, number) pairs that will form our final $\bm{x}$.

In step 1, we first collect all team/player/city names as all entities in the training set. We locate these entities in $\bm{y}$ and replace every multi-word entity name in $\bm{y}$ by an underscore-connected single token in order to simplify the task. These tokens are then recognized as entities. To find out the numbers, we simply invokes a Python module called text2num (Copyed from \href{https://github.com/exogen/text2num/blob/289745aebaf91e312fa8f8d86e04c17d7a3771af/text2num.py}{exogen's text2num.py}) which can recognize and convert every English numbers in $\bm{y}$ to digital numbers. Then all digital numbers are recognized.

In step 2, we enumerate every pair of (entity, number) found in step 1 and retrieve all candidate data records from the table. We simply iterate over the table and pick out those records whose entity name and score are exactly the same as the entity and number in our pair. However, there can be multiple records picked out which have different or even contradict labels due to the ambiguity. To reduce  wrong records retrieved, we add more rule constraints to filter out obviously wrong records. For instance, if the succeeding token of the number is `assist', then those records with labels related to `rebound' or `turnover' are obviously wrong. After this, there is still some redundancy in these records, though, we believe it is accurate enough to collect these records as $\bm{x}$. Finally, the entity names are also added to $\bm{x}$ so that the copy mechanism can directly copy the entity names from $\bm{x}$.

After obtained all $(\bm{x}, \bm{y})$ pairs of the training/validation/test sets, we now have to assign a $(\bm{x}_{e}, \bm{y}_{e})$ to each of them. The $(\bm{x}_{e}, \bm{y}_{e})$ is always from the training set, even for $(\bm{x}, \bm{y})$ in the validation/test sets.

The $(\bm{x}_{e}, \bm{y}_{e})$ should meet our requirement that $\bm{x}$ and $\bm{x}_{e}$ are analogous but not completely match in their labels.

\subsection{Restaurant Dataset}\label{sec:E2E creation}

The restaurant dataset is devised from the ~\citep{dusek2019e2e}, which consists of the review records containing
slots and values and the corresponding natural language. Note that one of the characteristics of the dataset is that the above records are non-aligned with the descriptive sentences, e.g., \emph{`customRating: 1\_out\_of\_5'} may be expressed as `low rating' in the descriptive sentence. This feature hinders us using BLEU to evaluate the exemplar preservation to some extent. Hence, we adopt some pre-processing to handle this problem as below.

To obtain our data, we first process each record into a tuple. In the similiar way with the NBA one, we replace the multi-word entity name with an underscore-connected  single token for simplicity. For some records e.g., `\emph{familyFriendly: no}', whose value cannot be aligned with the descriptive phrase, we directly replace the value `\emph{no}' with the phrase with the same meaning, i.e., `\emph{not\_family\_friendly}'. 

Then we need to retrieve a $(\bm{x}_{e}, \bm{y}_{e})$ from the training set. Unlike the NBA dataset, to make the dataset more challenging, we allow more different structures between $\bm{x}$  and $\bm{x}_{e}$. Concretely, the retrieved record $\bm{x}_{e}$ is designated to have the same number of tuples yet different data types with $\bm{x}$. We fix the number of unpaired data types to be 2. For those records without so many unpaired types, we set the number to be 1 or 0 as much as possible.

\section{Slot-filling Method}

 We present the hand-crafted rules of the slot-filling method for both two datasets. The first step is to match between $\bm{x}$, $\bm{x}_{e}$ and $\bm{y}_{e}$ with a large set of rules, and then roughly replace corresponding portions in $\bm{y}_{e}$ with those in $\bm{x}$. Specifically, we first build a mapping between the tuples of $\bm{x}$ and $\bm{x}_{e}$ through their data types, and then remove the content words in the exemplar sentence $\bm{y}_{e}$ to make a template. 
 Afterwards, we build a mapping between $\bm{x}_{e}$ and $\bm{y}_{e}$ through data values, types and indicative tokens (e.g., ``12 points'' in $\bm{y}_{e}$ indicates 12 is of type player points or team\_points). The two mappings connect $\bm{x}$ and $\bm{y}'$.
 Finally, these method fills in the slots with respective values in the record $\bm{x}$. enabling us to swap appropriate text in $\bm{y}_{e}$ to express content $\bm{x}$. 
    % Intuitively, Slot filling  method would perform well in preserving the style, as it merely replaces content related tokens without modifying other parts of the style sentence. However, this approach tends to leave out some new content and retain irrelevant information due to the different structures between $\bm{x}$ and $\bm{x}_{e}$.}

\section{Details for MAST}
The method is based on back-translation~\citep{sennrich2015improving} that first generates a target sentence $\bm{\hat{y}}$ conditioning on $(\bm{x}, \bm{y}_{e})$, and then treat it as the reference to reconstruct $\bm{y}_{e}$ conditioning on $(\bm{x}_{e}, \bm{\hat{y}})$. Auxiliary sentence $\bm{y}_{x}$ is used in an extra auto-encoding loss.

% is flexible for the diverse pattern of reference sentences.       
\section{ Details on the Excl.-old Metric and Bert Classifier}\label{sec:classifier}

In this section, we provide more details on evaluating the Excl.-old for the restaurant dataset. As mentioned above, there are different structures (i.e. data types) between the content records $\bm{x}$ and $\bm{x}_{e}$. Hence, the old content portions from the exemplar sentence $\bm{y}_{e}$ may appear in the generated sentence $\bm{\hat{y}}$ acting as Excl.-old. To evaluate the Excl.-old quantity, we firstly concatenate each element (i.e. the data tuple) of each record $\bm{x}$ with $\bm{\hat{y}}$ respectively and then determine whether the former appear in the latter via a reliable Bert classifier. Considering the third case in Table \ref{tab:exp-examples} as an example, we firstly concatenate each element of each record $\bm{x}$ with corresponding $\bm{\hat{y}}$ respectively, e.g., `\emph{name: Cocum | Cocum is a coffee\_shop that serves Italian food priced at \pounds20-25 . It is family\_friendly and it has a high customer rating .}' Then we use a binary Bert classifier to determine whether the generated sentence express the content tuple. The content fidelity denotes the percentage across all the new content tuples that appear in corresponding $\bm{\hat{y}}$. In contrast, the Excl.-old denote the overall percentage of the old content appearing in $\bm{\hat{y}}$ respectively.

Next we present the motivation and details of the Bert classifier. As mentioned in \ref{sec:E2E creation}, the restaurant dataset has the non-aligned characteristic, which  motivates us to use a Bert classifier to learn the semantic mappings. Note that the domain-specific dataset needs additional pre-training from the Bert checkpoint, which consumes a large amount of computation. Therefore, we train the classifier from scratch without using the pre-trained Bert model. 
\begin{table*}[t]
\hspace{-0.5cm}
%\centering
\small
\begin{tabular}{@{}r | l@{}}
\cmidrule[\heavyrulewidth]{1-2}
  {\bf Content Record}  &  
 \begin{tabular}{@{}l l l l l l l l l l l @{}}
 {\bf PLAYER}&{\bf PLAYER}&{\bf PTS} & {\bf FGM} & {\bf FGA} & {\bf FG3M} & {\bf FG3A} & {\bf FTM} & {\bf FTA} & {\bf AST} & {\bf MIN}\\
 Kemba\_Walker&Brian\_Roberts &18 & 6 & 10 & 1 & 3 & 5 & 5 & 3 &34
 \end{tabular} 
 \\ \cmidrule{1-2}
{\bf Exemplar Reference}  & Deron\_Williams \style{nearly matched} Turner \style{with a triple - double of his own , posting} 10 \style{points (} 2 - 6 \style{FG ,} 0 \style{-} 2 \style{3Pt ,}\\
&6 \style{-} 6 \style{FT ) }  , 9 rebounds and 10 \style{assists in} 34 \style{minutes .} \\[2pt] \cmidrule{1-2}
Slot filling  & Kemba\_Walker nearly matched Brian\_Roberts with a triple - double of his own , posting 18 points ( 6 - 10 FG , 1 - 3 \\
&3Pt , 5 - 5 FT ) , \error{and 9 rebounds} and 3 assists in 34 minutes . \\[2pt] \cmidrule{1-2}
AdvST & Kemba\_Walker nearly matched Brian\_Roberts with a triple - double of his own , posting 18 points ( 6 - 10 FG , 1 - 3\\
&3Pt , 5 - 5 FT ) , \error{and 5 assists} and 3 assists in \error{38} minutes . \\[2pt] \cmidrule{1-2}
Ours w/o Cover. & Kemba\_Walker \error{[...]} had a triple - double of his own , posting 18 points ( 6 - 10 FG , 1 - 3 3Pt , 5 - 5 FT ) , and 3 \\
& assists in 34 minutes .\\[2pt] \cmidrule{1-2}
Ours & Kemba\_Walker \style{nearly matched} Brian\_Roberts \style{with a triple - double of his own , posting} 18 \style{points (} 6 \style{-} 10 \style{ FG ,} 1 \style{-} 3 \\
&\style{3Pt ,} 4 \style{-} 4 \style{FT ) and } 3 \style{assists} in 34 \style{minutes .} \\[2pt]

%\cmidrule[\heavyrulewidth]{1-2}
\cmidrule{1-2}\morecmidrules\cmidrule{1-2}
  {\bf Content Record} &  
 \begin{tabular}{@{}l l l l l l l l l @{}}
 {\bf Name} & {\bf EatType} & {\bf Food} & {\bf PriceRange} & {\bf CustomRating} & {\bf Area} & {\bf FamilyFriendly} \\
  Fitzbillies & coffee\_shop & French & \pounds20-25 & high & riverside & family\_friendly 
 \end{tabular} 
 \\ \cmidrule{1-2}
{\bf Exemplar Reference} & The\_Punter \style{is a} family\_friendly coffee\_shop \style{in the} less\_than\_\pounds20 \style{range} . \style{They serve} Japanese \style{food and can be found} \\
& \style{near} Café\_Sicilia . \style{They have a} low \style{customer rating} .\\[2pt] \cmidrule{1-2}
Slot filling  & Fitzbillies is a family\_friendly coffee\_shop in the \pounds20-25 range . They serve French food and can be found near\\
&\error{Café\_Sicilia} . They have a high customer rating . \\[2pt] \cmidrule{1-2}
AdvST & \error{The\_Punter} \error{[...]} is a family\_friendly coffee\_shop \error{[...]} in the riverside area . They serve French food and can be found\\
&near riverside . They have a high customer rating . \\[2pt] \cmidrule{1-2}
Ours w/o Cover. & Fitzbillies is a family\_friendly coffee\_shop in the \pounds20-25 range . They serve French food and can be found near\\
&\error{Café\_Sicilia} . They have a high customer rating . \\[2pt] \cmidrule{1-2}
Ours &  Fitzbillies \style{is a} family\_friendly coffee\_shop \style{in the} \pounds20-25 \style{range} . \style{They serve} French \style{food and can be found} \style{ near}\\
&riverside . \style{They have a} high \style{customer rating} . \\[2pt]
\cmidrule[\heavyrulewidth]{1-2}
\end{tabular}
\vspace{-8pt}
\caption{Example Outputs by Different Models. Text of erroneous content is highlighted in \error{red}, where \error{[...]} indicates desired content is missing. Text portions in the exemplar sentences and the generated sentences by our model that fulfill the stylistic characteristics are highlighted in \style{blue}.}
\label{tab:exp-examples}
\end{table*}
\begin{table*}%[!b]
\hspace{-0.5cm}
%\centering
\small
\begin{tabular}{@{}r | l@{}}
\cmidrule[\heavyrulewidth]{1-2}
 {\bf  Content Record}  &  
 \begin{tabular}{@{}l l l l l l l l l @{}}
 {\bf PLAYER} & {\bf FGM} & {\bf FGA} & {\bf FG3M} & {\bf FG3A} & {\bf PTS} & {\bf MIN} \\
 Johnson & 6 & 9 & 4 & 5 & 16 & 20
 \end{tabular} 
 \\ \cmidrule{1-2}
{\bf Exemplar Reference}  & Rudy\_Gay led his team , with 16 points ( 6 - 12 FG , 4 - 6 3Pt ) and 7 rebounds in 27 minutes . \\[2pt] \cmidrule{1-2}
Ours & Johnson led his team , with \error{9} points ( 6 - 9 FG , 4 - 5 3Pt ) \error{and 16 rebounds } in 20 minutes . \\[2pt] \cmidrule{1-2}
Desired Output & Johnson led his team , with 16 points ( 6 - 9 FG , 4 - 5 3Pt ) in 20 minutes . \\[2pt]
\cmidrule{1-2}\morecmidrules\cmidrule{1-2}
 {\bf  Content Record}  &  
 \begin{tabular}{@{}l l l l l l l l l @{}}
 {\bf Name} & {\bf Food} & {\bf PriceRange} & {\bf FamilyFriendly} & {\bf Near} \\
The\_Golden\_Curry & Japanese & moderate & family\_friendly & The\_Bakers
 \end{tabular} 
 \\ \cmidrule{1-2}
{\bf Exemplar Reference} &The\_Phoenix is a restaurant in riverside that serves moderately priced French food . It has a customer rating\\
&of 1\_out\_of\_5 .\\[2pt] \cmidrule{1-2}
Ours & The\_Golden\_Curry is a restaurant \error{in} The\_Bakers that serves moderately priced Japanese food . It has a \error{customer}\\
&\error{ rating} of family\_friendly .\\[2pt] \cmidrule{1-2}
Desired Output & The\_Golden\_Curry is a restaurant near The\_Bakers that serves moderately priced Japanese food. It has an \\
& atmosphere of family\_friendly . \\[2pt]
\cmidrule{1-2}\morecmidrules\cmidrule{1-2}
  {\bf Content Record}  &  
 \begin{tabular}{@{}l l l l l l l l l @{}}
 {\bf Name} & {\bf EatType} & {\bf Food} & {\bf PriceRange} & {\bf CustomerRating} & {\bf Area}   \\
 The\_Golden\_Palace & coffee\_shop & Indian & moderate & 1\_out\_of\_5 & riverside  \\
 \end{tabular} 
 \\ \cmidrule{1-2}
{\bf Exemplar Reference}  & The Strada restaurant providing sushi is located near Rainbow\_Vegetarian\_Café . It is medium priced and has a\\
& 3\_out\_of\_5 star rating . \\[2pt] \cmidrule{1-2}
Ours & The\_Golden\_Palace \error{restaurant} providing \error{sushi} is located near riverside . It is moderate priced and has a\\
& 1\_out\_of\_5 star rating . \\[2pt] \cmidrule{1-2}
Desired Output &The\_Golden\_Palace coffee\_shop providing Indian food is located near riverside . It is moderate priced and has\\
&a 1\_out\_of\_5 star rating .\\[2pt]
\cmidrule[\heavyrulewidth]{1-2}
\end{tabular}
\vspace{-8pt}
\caption{Example Erroneous Outputs. Text of erroneous content is highlighted in \error{red}. Missing content is denoted with \error{[...]}. We also show the desired correct outputs.}
% In the  first example, the model was confused by the data types; while in the second example, the model fails to understand there is only one team in the content record $\bm{x}$ .}
\label{tab:examples-error}
\end{table*}

The process of constructing the dataset for training is as follows. Firstly, we concatenate each element of $\bm{x}$ with $\bm{y}_{x}$ and label it to be the positive sample. Then, to obtain the negative sample, we concatenate each element of $\bm{x}_{e}$ with $\bm{y}_{x}$ and 
label it to be the negative sample. The positive-negative sample ratio is set to be 1 : 3. The train/dev/test set is devised from the corresponding sets of the restaurant reviews dataset. The bert classifier achieves 94\% accuracy on the test set, which is reliable to evaluate the content fidelity and Excl.-old.

\section{ Additional Qualitative Examples}\label{sec:addtional-examples}

Table~\ref{tab:exp-examples-error} shows a failure case by the proposed model along with ideal outputs. Despite the superior performance over other approaches, the model can get confused about rare content records (e.g., ``88'' is the free-throw percentage). It is desirable to further improve the modeling of both content and style to better capture the semantics and fulfill better generation results.

\begin{table*}[!h]
\centering
\small
\begin{spacing}{0.7}
\begin{tabular}{@{}r | p{13.8cm}@{}}
\cmidrule[\heavyrulewidth]{1-2}
  {\bf Content Record} &  
 \begin{tabular}{@{}l l l l l l l l l @{}}
 {\bf TEAM} & {\bf TEAM-FG\_PCT} & {\bf TEAM-FG3\_PCT} & {\bf TEAM-FT\_PCT}   \\
 Kings & 55 & 50 & 88 \\
 \end{tabular} 
 \\ \cmidrule{1-2}
{\bf Exemplar} & Portland allowed the Spurs to shoot a scorching 50 percent from field and 60 percent from beyond arc. \\[2pt] \cmidrule{1-2}
Ours & The Kings \error{allowed the Kings to} shoot a scorching 55 percent from field and \error{88} percent from beyond arc \error{[...]}. \\[2pt] \cmidrule{1-2}
Desired Output & Kings shot a scorching 55 percent from field, 50 percent from beyond the arc, and 88 percent from free throw line .\\[2pt]
\cmidrule[\heavyrulewidth]{1-2}
\end{tabular}
\end{spacing}
\vspace{-10pt}
\caption{Example erroneous outputs. Text of erroneous content is highlighted in \error{red}. Missing content is denoted with \error{[...]}. We also show the desired correct outputs.
In this example, the model fails to understand there is only one team in the content record $\bm{x}$ and the number 88 is the free-throw percentage.}
\label{tab:exp-examples-error}
\end{table*}
\bibliography{emnlp2020}
\bibliographystyle{acl_natbib}